\documentclass[fleqn,10pt]{wlscirep}
\usepackage[utf8]{inputenc}
\usepackage[T1]{fontenc}
\usepackage{tcolorbox}
\usepackage{xcolor}
\usepackage{enumitem}
\title{Correspondence of high-dimensional emotion structures elicited by video clips between humans and Multimodal LLMs}

\author[1]{Haruka Asanuma}
\author[4, 5]{Naoko Koide-Majima}
\author[2]{Ken Nakamura}
\author[3]{Takato Horii}
\author[4, 5, 6]{Shinji Nishimoto}
\author[1, *]{Masafumi Oizumi}
\affil[1]{The University of Tokyo, Graduate School of Arts and Sciences, Tokyo, 153-8902, Japan}
\affil[2]{The University of Tokyo, Faculty of Engineering, Tokyo, 113-8656, Japan}
\affil[3]{The University of Osaka, Graduate School of Engineering Science, Osaka, 565-0871, Japan}
\affil[4]{Center for Information and Neural Networks (CiNet), National Institute of Information and Communications Technology, Osaka, 565-0871, Japan}
\affil[5]{The University of Osaka, Graduate School of Frontier Biosciences, Osaka, 565-0871, Japan}
\affil[6]{The University of Osaka, Graduate School of Medicine, Osaka, 565-0871, Japan}

\affil[*]{c-oizumi@g.ecc.u-tokyo.ac.jp}


\begin{abstract}
Recent studies have revealed that human emotions exhibit a high-dimensional, complex structure. A full capturing of this complexity requires new approaches, as conventional models that disregard high dimensionality risk overlooking key nuances of human emotions. Here, we examined the extent to which the latest generation of rapidly evolving Multimodal Large Language Models (MLLMs) capture these high-dimensional, intricate emotion structures, including capabilities and limitations. Specifically, we compared self-reported emotion ratings from participants watching videos with model-generated estimates (e.g., Gemini or GPT). We evaluated performance not only at the individual video level but also from emotion structures that account for inter-video relationships. At the level of simple correlation between emotion structures, our results demonstrated strong similarity between human and model-inferred emotion structures. To further explore whether the similarity between humans and models is at the signle-item level or the coarse-categorical level, we applied Gromov–Wasserstein Optimal Transport. We found that although performance was not necessarily high at the strict, single-item level, performance across video categories that elicit similar emotions was substantial, indicating that the model could infer human emotional experiences at the category level. Our results suggest that current state-of-the-art MLLMs broadly capture the complex high-dimensional emotion structures at the category level, as well as their apparent limitations in accurately capturing entire structures at the single-item level.
\end{abstract}
\begin{document}
\flushbottom
\maketitle
\thispagestyle{empty}
\noindent
\section*{Introduction}
Recent studies have revealed that human emotions possess an exceptionally complex and high-dimensional structure\cite{Cowen2017-lx,Koide-Majima2020-kc}. Traditionally, emotion research has focused on simplifying emotions into lower-dimensional models to make them more manageable. For instance, basic emotion theory \cite{Ekman1971-it} proposes that human emotions can be categorized into six fundamental types, while the dimensional approach \cite{Russell1980-tm, Russell1999-ej} maps emotions onto a two-dimensional space defined by arousal (active–inactive) and valence (positive–negative). These models have been widely applied in fields such as facial expression recognition and continue to exert significant influence in psychology and artificial intelligence\cite{Maithri2022-zq, Breazeal2004-ja}. 
However, recent studies employing data-driven approaches \cite{Cowen2017-lx,Koide-Majima2020-kc, Keltner2019-uw} suggest that human emotions cannot be fully captured within such low-dimensional frameworks, but instead, exhibit a more complex, high-dimensional structure. For example, by using self-reports while watching videos, Cowen \& Keltner \cite{Cowen2017-lx} identified 27 distinct emotional dimensions based on large-scale self-reported data and Koide-Majima et al. \cite{Koide-Majima2020-kc} identified 18–36 brain-correlated emotion dimensions per participant using fMRI and 80 emotion categories, highlighting the subtle differences and intricate interrelationships among emotions. This evidence indicates that accurate modeling and understanding of human emotions requires a more sophisticated approach which explicitly accounts for the high-dimensional nature of emotional experiences.

Against this backdrop, an emerging approach is to leverage large language models (LLMs), particularly multimodal LLMs (MLLMs), to understanding human emotions. MLLMs, integrating capabilities for processing text, images, and audio into LLMs, have rapidly advanced over recent years, and now have the ability to process multiple modalities, not only text but also images and audio. They have already demonstrated high performance in emotion inference tasks based on external expressions, such as facial expression recognition \cite{Foteinopoulou2024-kt, Lian2024-fb}, text-based sentiment analysis \cite{Liang2022-mx, Zeng2023-mq, Zhang2025-au} and tri-modal emotion recognition from speech, textual content, and facial expressions\cite{Li2024-ng, Cheng2024-cy}. If these MLLMs can move beyond such limitations and accurately replicate the complexity of human affective responses, they might be able to serve as a valuable new tool for emotion research.

However, the question of whether MLLMs can accurately infer the high-dimensional structures of emotions and also the emotions experienced internally by humans, for example while watching videos, has not been answered, and doing so is challenging. This is because these tasks require a multi-step inference process that goes beyond simple feature extraction\cite{Fei2024-fg}. Specifically, predicting how a person will feel while watching a video involves two key steps: first, accurately recognizing what is depicted in the video, and second, reasoning about how the viewer will respond emotionally to it. This process is inherently complex, and depends on multiple factors such as narrative context and the viewer's prior knowledge. It is therefore fundamentally different from the simple analysis of expressed emotions. While recent studies suggest that MLLMs have the potential to infer subjective sensory experiences such as color perception and auditory pitch \cite{Kawakita2024-nd, Marjieh2024-br}, their ability to accurately estimate more abstract, context-dependent emotions remains uncertain.

\begin{figure}[!h]
    \centering
    \includegraphics[width = 0.9\linewidth]{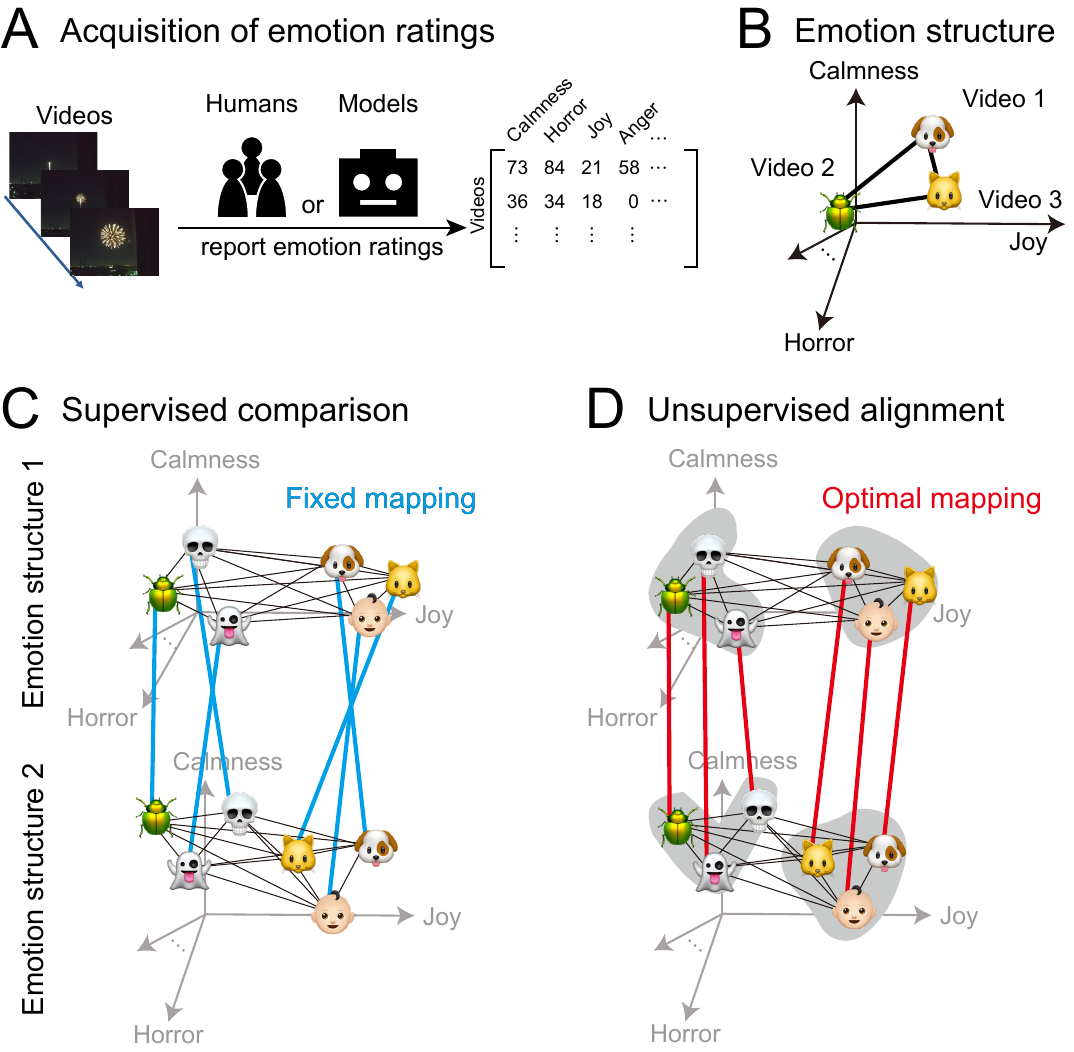}
     \caption{Overview of the analytical framework for comparing similarity structures of emotions across two domains (e.g., humans vs. model).  \textbf{A: Acquisition of emotion ratings.} Participants and models watch a series of video clips and report emotion ratings on multiple dimensions, such as calmness, joy, horror, anger. The elements of the matrix represent the intensity of each emotion category for each video reported by participants or models. \textbf{B: Emotion structures.} Each video's emotion ratings, as reported by humans and models, are represented as points in a multidimensional space to illustrate the relational structure of emotions (emotion structure). The points corresponding to videos that evoke similar emotional responses, such as Video 1 (dog) and Video 3 (cat) associated with joy, are positioned closer together, while videos eliciting distinct emotions, such as Video 2 (insect) associated with horror, are placed further apart. Dissimilarity between videos is represented by distance, namely the black lines between points. \textbf{C: Supervised comparison of emotion structures.}  Supervised comparison of emotion structures between two domains based on fixed mapping between the same videos, which is represented by blue lines. \textbf{D: Unsupervised comparison of emotion structures.} A conceptual illustration of  unsupervised comparison based on Gromov-Wasserstein Optimal Transport (GWOT), which searches for optimal mappings based solely on internal relations (RDMs). The optimal mappings are shown as red lines. In the figure, groups of videos that evoke similar emotions (categories) are surrounded by gray outlines. In this case, the mappings are categorical but not exact at the fine-item level, e.g., ghost is mapped to skull and dog is mapped to cat, but these are appropriately paired within the same category.
    }
    \label{fig:overview}
\end{figure}

In this study, we investigated the extent to which current MLLMs can accurately predict the emotions that people experience when watching videos (Figure \ref{fig:overview}A). To this end, we used a dataset from previous studies in which participants reported emotion ratings they experienced while watching video clips \cite{Cowen2017-lx,Koide-Majima2020-kc}. We then instructed MLLMs, including Gemini\cite{Pichai2023-qi} and GPT\cite{Open2022-fk}, to report emotion ratings for these videos (Figure \ref{fig:overview}A), and assessed how well the models' predicted emotions matched the human ratings.

In evaluating these ratings from humans and models, we not only examined agreement in emotion ratings for each video, but also focused on patterns and relationships across multiple videos — that is, the emotion structure (Figure \ref{fig:overview}B). Emotion structure refers to the relational structures of emotions elicited by videos. In this study, relationships are specifically similarity or dissimilarity among the multidimensional emotional responses evoked by different videos. Figure \ref{fig:overview}B visually represents these relationships by mapping each video's emotion ratings into a multidimensional space. For instance, Video 1 (dog) and Video 3 (cat) are both associated with joy and are therefore located close to each other in the space. In contrast, Video 2 (insect), which evokes the emotion horror is positioned further away. The spatial distances and distribution patterns among these videos collectively represent the emotion structure.

The reason we focused on comparing emotion structures across multiple videos rather than performing direct, one-to-one comparisons of emotion ratings is that people and models can differ in how they interpret and use emotion terms. For example, even a seemingly straightforward emotion like ``joy'' may be used differently for emotion ratings by different people or computational models. Indeed, previous research has shown that emotion ratings are influenced by individual cognitive tendencies and model-specific biases, leading to considerable variability among humans \cite{DiGirolamo2023-ti, Barrett2017-vn} and among computational models \cite{Vaiani2024-si}. In contrast, focusing on emotion structures allows us to abstract away from differences in the specific use of emotion terms and instead concentrate on relative similarity or dissimilarity among emotional responses. This structural approach primarily evaluates whether relationships between emotional content — such as the distinction between ``joy'' and ``horror'' videos — are consistently represented, irrespective of the exact emotion terms used to express the emotions elicited by these videos. Thus, this method enables us to assess how accurately models capture human emotional recognition patterns based on relational structures, and minimizes confounding related to terminological variability.

In this study, we explored similarities and differences in emotion similarity structures by applying two complementary comparison strategies, supervised and unsupervised, following the methodologies proposed in previous studies \cite{Takeda2025-na, Takeda2025-li, Kawakita2025-kt}. As shown schematically in Figures \ref{fig:overview}C and \ref{fig:overview}D, the supervised approach assumes a fixed one-to-one correspondence between emotional responses to the same videos. Conventional representational similarity analysis (RSA) takes this approach and then evaluates the similarity between structures by computing the correlation between representational dissimilarity matrices (RDMs) of different domains (e.g., humans and models). In contrast, the unsupervised approach, specifically Gromov-Wasserstein Optimal Transport (GWOT), which we use in this study, searches for the mapping that best aligns the two structures purely from their internal relational geometry, allowing different elements to be optimally paired. For example, in Figure \ref{fig:overview}D, ghost aligns with skull and dog with cat, resulting in categorical rather than item-level agreement. When such mappings occur, we may conclude that the two structures are ``categorically'' matched, i.e., joyful videos – dog, cat, baby – are correctly mapped to the same category of images, and horror images – ghost, skull, insect – are also correctly mapped to the same category of images, but this mapping is not a fine-item-level one-to-one mapping. The juxtaposition of these methods allows us to distinguish fine item-level alignment from coarser category-level alignment (see previous studies\cite{Takeda2025-na, Takeda2025-li} for details), thus providing a comprehensive assessment of how well model-predicted emotion structures match human judgments, from individual videos to higher-order abstract organizations.

\section*{Results}
\subsection*{Datasets and analysis methods}
In the following analysis to investigate the structures of human emotions, we used two datasets from previous studies: Koide-Majima et al. (2020) and Cowen and Keltner (2017). Both datasets are subjective emotion ratings of human participants while watching short video clips (see Methods for the details). 

The following results are organized as follows. First, before evaluating the ability of the MLLMs, we first examined whether there were common emotion structures across human participants. The degree of commonality between human participants can be considered as an approximate upper-bound of the degree of commonality between actual human emotion structures and that inferred by MLLMs. For this we used data from Koide‐Majima et al. (2020) only, due to the limited data availability of Cowen and Keltner (2017). After that, we evaluated the degree of similarity between emotion structures of humans and those inferred by MLLMs by using both datasets. 

To evaluate the degree of commonality in emotion structures between human participants or between humans and models, we performed two types of analyses on each dataset. The first analysis focused on the similarity of emotion reports for individual videos, while the second analysis considered the overall similarity structure of emotions elicited by different videos, not restricted only within each video. For the second analysis of the emotion similarity structure, we employed two metrics: the correlation between emotion similarity structures, known as the conventional Representational Similarity Analysis (RSA), which assumes a correspondence between videos; and Gromov–Wasserstein Optimal Transport (GWOT), which does not assume such a correspondence (see Methods for details). The importance of using GWOT is to find the optimal mapping between emotion structures without video labels, and then based on the optimal mapping, to check whether the same videos correspond to each other in the emotion structures of humans and models. We evaluate the matching rate of one-to-one correspondence of each video to assess whether the structures are matched at the fine-item level. Since it is possible that the structures are not matched at the fine-item level, but are matched at the coarse-category level, we also evaluated the matching rate of coarse-category correspondence of a group of videos. 

\subsection*{Commonality of emotion structure between different participant groups}
In this section, we examine the extent to which different groups of participants shared a common emotion structure, using data from Koide‐Majima et al. (2020) \cite{Koide-Majima2020-kc} (Figure \ref{fig:overview}A). In the dataset, participants watched 550 short video clips and continuously rated the intensity of their emotional responses on a scale from 0 to 100. Each participant reported one or two emotions per video, resulting in a total of 80 emotion categories (see Methods for details). Overall, each emotion category was reported by four participants.

To evaluate the consistency of emotion structures among participants, we randomly split the emotion rating data for each emotion category into two groups. Each group contained ratings from two participants per emotion category, and each group was treated as a pseudo-participant by aggregating and averaging the ratings across individuals. 
It should be noted, however, that because each participant could report up to two different emotions, it is possible that ratings from the same participant may appear in different groups, and thus there is an overlap of participants involved in each group.
However, we ensured that there was no overlap within the same emotion category — that is, no individual participant's data appeared in both groups for the same emotion. This group assignment method guarantees that similarity between groups is not artificially inflated.

\subsubsection*{Similarity between emotion ratings for each video}
\begin{figure}[ht]
    \centering
    \includegraphics[width = 0.5\linewidth]{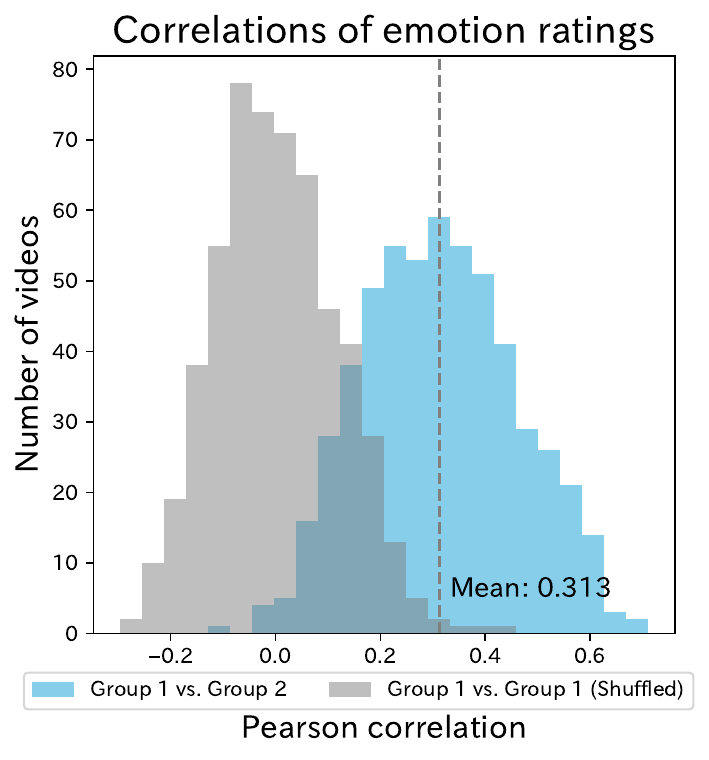}
    \caption{
    Histogram of the Pearson correlation for each video clip between human ratings in the Koide-Majima et al. dataset. The blue histogram represents the distribution of the correlation between the ratings of Participant group 1 and group 2 participants for each video, and the gray histogram represents the distribution of the correlation between the Participant group 1 ratings and the shuffled Participant group 1 ratings, which served as the null distribution. The dashed line represents the mean of the correlation, 0.313, between the ratings of Participant group 1 and group 2 (blue histogram). 
    }
    \label{fig:nishimoto_human_hist}
\end{figure}

The analysis of similarity between the emotion ratings of the two participant groups for each video revealed that the ratings are relatively consistent between participant groups (Figure \ref{fig:nishimoto_human_hist}). The blue histogram represents the distribution of Pearson correlation coefficients between the two groups (Participant group 1 and group 2) for each video, with the mean correlation value of 0.313 shown by the dashed line. To assess statistical significance, we also estimated correlation values at the chance level by computing Pearson correlation coefficients between the responses of Participant group 1 and the shuffled responses of Participant group 1. The gray histogram shows the distribution obtained by a one-time shuffling of the responses (see Methods for details). The Cohen's D between the distribution of correlation coefficients (blue) and the null distribution by one-time shuffled data (gray) was 2.33. Cohen's D computed from 1,000 shuffles was 2.64, demonstrating a statistically significant difference between the original distribution and the null distribution. Although the correlation values are statistically significant, the average value of 0.313 is at a moderate level and not necessarily high, indicating that a significant level of individual differences in emotion ratings exists between participant groups at the level of each individual video.

\subsubsection*{Similarity between similarity structures of evoked emotions from all videos}
We performed two analyses, conventional RSA and GWOT, to evaluate the similarity between the entire similarity structures formed by the emotion ratings of all videos between the participant groups.

\begin{figure}
    \centering
    \includegraphics[width = \linewidth]{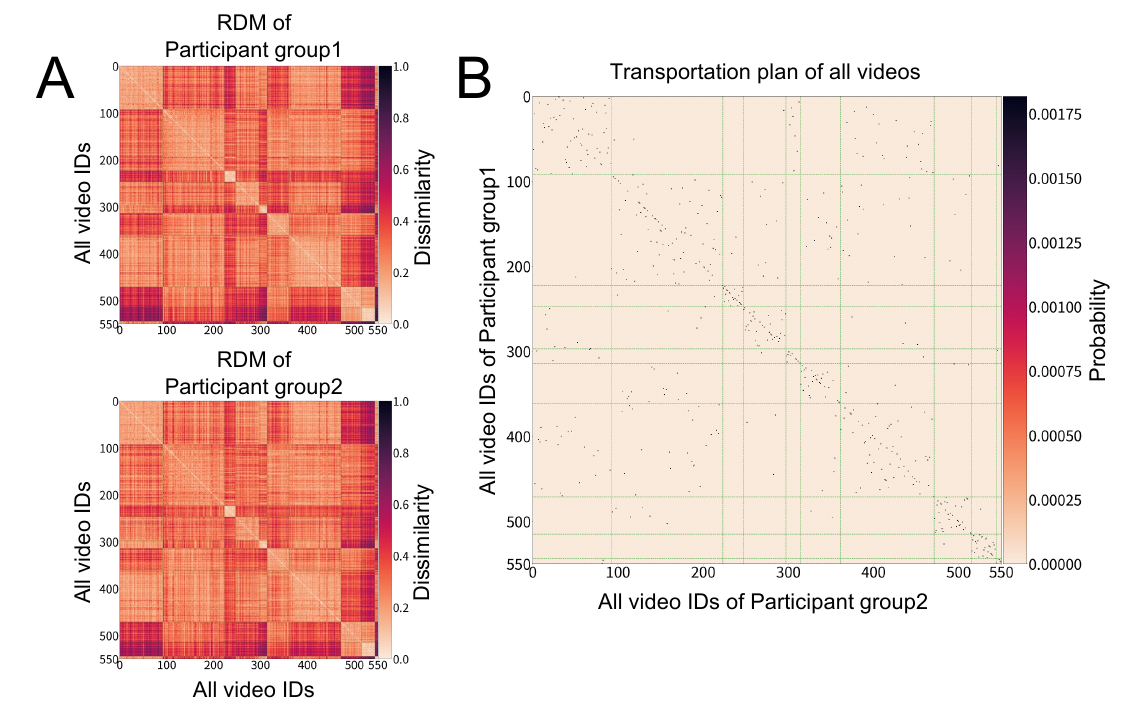}
    \caption{
    Unsupervised comparison of the similarity structures for all videos in the Koide-Majima et al. dataset between Participant group 1 and group 2 based on Gromov-Wasserstein Optimal Transport (GWOT). A: Representation Dissimilarity Matrices (RDMs) of Participant group 1 and group 2. The elements of the RDMs represent the dissimilarity between the emotion ratings of the videos, quantified by cosine similarity.  B: Optimal transportation plan obtained by GWOT between the RDMs of Participant group 1 and group 2. Green lines represent the category boundaries of the videos.  
    }
    \label{fig:nishimoto_human_RDM_GW}
\end{figure}

\paragraph{Correlation between similarity structures}
To evaluate the similarity of the emotion structures, we first calculated the correlation between the similarity structures of the emotion reports of all videos across participant groups. Figure \ref{fig:nishimoto_human_RDM_GW}A shows the representation dissimilarity matrices (RDMs) for Participant groups 1 (left) and 2 (right). To create the RDMs, we used cosine similarity to measure the similarity between the emotion ratings of the videos. We found that the correlation coefficient between the RDMs was markedly high, 0.859, which significantly exceeds the chance level, estimated via 1,000 shuffles with a 95\% percentile interval ($[-0.000509, 0.00108]$, Table \ref{tab:nishimoto_Gemini}). 
The level of similarity at the level of the representational similarity structures is significantly higher than the level of similarity at the level of each individual video (0.31 on average) evaluated in the previous section. 

\paragraph{GWOT: One-to-one matching}
Next, by performing GWOT, we found that the one-to-one matching rate of the optimal transportation plan is also high (see Figure \ref{fig:nishimoto_human_RDM_GW}). 
As shown in the optimal transportation plan (Figure \ref{fig:nishimoto_human_RDM_GW}B), we can observe many non-zero
values in the diagonal elements. To be precise, the matching rate, i.e., percentage of non-zero diagonal elements (see Methods for details), is 16.36\%, which significantly exceeds both the theoretical chance level (0.182\%) and the empirically measured chance level, estimated via 10 shuffles with a 95\% percentile interval ($[1.01\%, 1.23\%]$, Table \ref{tab:nishimoto_Gemini}), as shown in Table \ref{tab:nishimoto_Gemini}. 
This finding means that even at the fine-item level and under this strict unsupervised alignment condition, the emotion structures between the participant groups have some degree of commonality, i.e., the same videos are unsupervisedly mapped at 16.36\%. This level of agreement can be treated as a rough estimate of the upper bound of agreement between humans and models, which will be done in subsequent sections.

\paragraph{GWOT: category matching}
Finally, by evaluating the matching rate of the optimal transportation plan obtained by GWOT at the coarse-category level, we found that the category matching rate is also high (Table \ref{tab:nishimoto_Gemini} and Figure \ref{fig:nishimoto_human_RDM_GW}). For analysis of category matching, we classified the videos into 10 categories via hierarchical
clustering of the participants' emotion reports (see Method in details). In Figure \ref{fig:nishimoto_human_RDM_GW}, the rows are sorted according to the hierarchical
clustering results, and the green lines indicate category boundaries. 
The non-zero values in the optimal transportation plan (Figure \ref{fig:nishimoto_human_RDM_GW}) concentrate along the diagonally outlined boxes (green lines), meaning a high degree of category alignment – 66.18\% – which is significantly higher than the empirical chance level ($[17.3, 19.4\%]$, Table \ref{tab:nishimoto_Gemini}).

This result means that a common emotion structure between different participant groups is also present at the coarse-category level. Similarly to the case with one-to-one matching, this level of agreement can be treated as a rough estimate of the upper bound of agreement between humans and models.

\subsection*{Evaluation of MLLM's estimation of human emotion structure for data from Koide-Majima et al. (2020)}
To assess whether MLLM can infer the high-dimensional similarity structure of human emotions, we compared the similarity structure of emotion ratings obtained by MLLM with that of humans. We specifically used Gemini for several criteria (see Methods, `Selection of Multimodal LLMs', for details). 
Using the same 550 videos employed in human psychology experiments, we obtained Gemini's response to the estimated emotion intensities for each video based on the prompt described in the Methods section.

\subsubsection*{Similarity between emotion ratings on each video}
\begin{figure}[t]
    \centering
    \includegraphics[width = 0.5\linewidth]{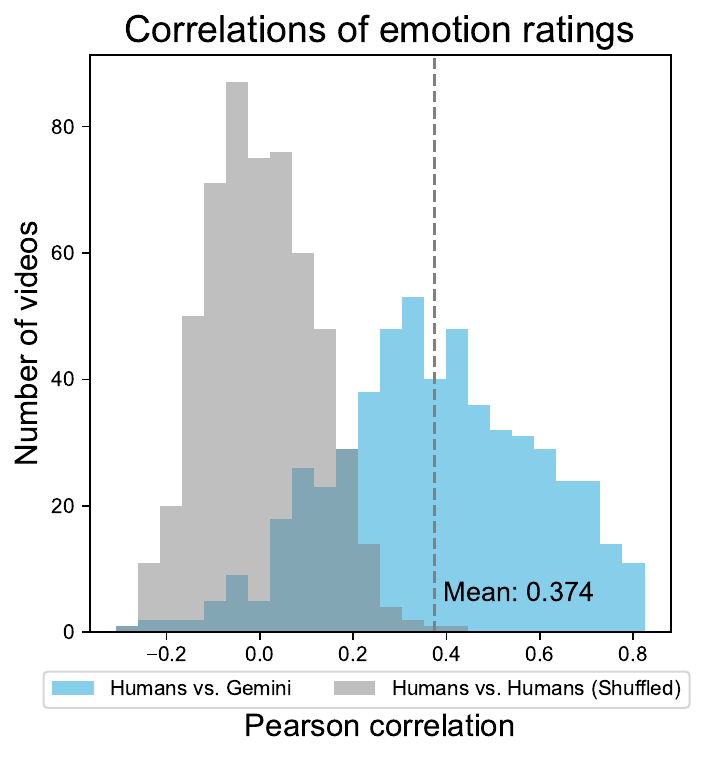}
    \caption{
    The histograms show the Pearson correlation of each video clip between the human ratings and Gemini's estimation in the Koide-Majima et al. dataset. The blue histogram represents the distribution of the correlation between the human ratings and the Gemini's estimation for each video, and the gray histogram represents the distribution of the correlation between the human ratings and the shuffled human ratings, which served as the null distribution. The dashed line represents the mean of the correlation, 0.374, between the human ratings and the Gemini's estimation (the blue histogram). 
    }
    \label{fig:nishimoto_human_Gemini_hist}
\end{figure}

Analysis based on the correlation calculated for each video revealed that Gemini's emotion estimates exhibit a degree of agreement with human ratings that is significantly higher than the chance level (Figure \ref{fig:nishimoto_human_Gemini_hist}). Specifically, the blue histogram shows the distribution of Pearson correlation coefficients between Gemini's estimates and the participant ratings for each video, while the gray histogram represents the distribution of correlations obtained after shuffling the human ratings. The dashed line indicates the mean correlation of the original data (0.374), and Cohen's D computed from 1,000 shuffles was 3.23, demonstrating a substantial difference between the original distribution and the null distribution. Furthermore, when comparing the distribution of correlation between Gemini's estimates and the participant ratings with the correlation observed between different participant groups (Figures \ref{fig:nishimoto_human_hist} and \ref{fig:nishimoto_human_Gemini_hist}), the correlation between Gemini and the participants was found to be slightly higher than that observed among the participant groups. These results ensure that – at least at the level of emotion ratings for each video – Gemini can provide human-like emotion ratings with variability comparable to the variability within human participants. 

\paragraph{Well-estimated and Poorly-estimated Videos}
To gain insight into possible reasons why some videos yielded higher similarity and others did not, we examined the content of the top 25 and bottom 25 ranked videos in terms of Pearson correlation. We selected five representative videos that were particularly typical in content, and summarized them in Table \ref{tab:contents_best_worst_of_nishimoto}.

By examining Table \ref{tab:contents_best_worst_of_nishimoto}, for videos that can evoke a strong emotional response from just a single frame, Gemini appears to estimate emotions with high accuracy. Specifically, in 273.mp4 and 477.mp4, which feature infants or pets (e.g., dogs or cats), the model's predictions of emotions such as ``cuteness'' and ``love'' closely matched the human participants' reports of ``adorableness'' and ``calmness.'' Likewise, in 048.mp4 and 527.mp4, which depict a dark setting with a black-haired woman wearing a white dress, even a single frame can strongly evoke ``fear,'' and the model accurately reproduced participants' reports of ``fear.'' Additionally, 024.mp4 shows a scene where surgical scissors and forceps are used to make an incision in the eyelid, and the model most strongly predicted the emotion ``empathic pain.'' Such videos are presumed to be the type where visual features alone nearly fully determine the emotion, thereby requiring minimal contextual or narrative elements, which in turn likely facilitates the model's accurate representation of the emotion structure.

By contrast, an analysis of the bottom five videos with the lowest correlation revealed two main reasons why Gemini struggled to accurately estimate emotions.

First, videos that require contextual interpretation tended to pose difficulties for the model. In particular, when a video's emotional meaning could shift dramatically depending on the context, the model often relied too heavily on surface-level visual features and produced incorrect estimates.
For example, in 188.mp4, a man is seen kissing a pregnant woman on the neck, followed by a scene in which he forcefully pushes her against a wall multiple times. While a single frame might appear romantic, the full video conveys an impression of sexual violence and threat, evoking discomfort and fear. In fact, whereas participants reported negative emotions, the model predicted high intensities of ``romantic'' and ``sexy.'' This shows that the model made its inference based solely on the visual feature of ``a couple kissing,'' and failed to recognize the contextual cues that indicate the woman was in danger.

Second, in short videos that provide insufficient information for even human viewers to comprehend the narrative and evoke appropriate emotions, the model likewise exhibited a decline in emotion prediction accuracy.
For instance, 292.mp4 is part of a movie trailer that depicts a man and woman in a decaying urban setting. However, based on the brief video alone, it is difficult to determine whether the story is about romance, a runaway, or something else entirely. Even human viewers may feel unsure about which emotion to report, and such ambiguity can lead to inconsistencies in participants' responses and lower prediction accuracy for the model.

Taken together, these findings show that while MLLMs demonstrate high accuracy when emotions can be inferred directly from visual features, their performance remains limited in situations that require contextual understanding.

\begin{table}[ht]
    \centering
    \begin{tabular}{c|p{6.5cm}||c|p{6.5cm}}
        \multicolumn{2}{c||}{Well estimated videos} & \multicolumn{2}{c}{Badly estimated videos} \\ \hline
        Video ID & Contents & Video ID & Contents \\ \hline
        048 & A long-haired woman in a white dress moves eerily and makes strange sounds as she approaches.  & 010 & A child who has a strained relationship with her father is being bullied at school. \\ 
        024 & The eyelid is being incised using surgical scissors and forceps. & 188&A man violently pushes a pregnant woman against a wall and forcibly kisses her.
        \\
        273 & A small cat being loved by its owner.  & 275 & A comedy movie trailer with many subtitles.\\ 
        477 & Two small dogs play together.& 
        292 & A movie trailer featuring a man and a woman living in a decaying city.
         \\
        527 & A creepy woman with long black hair in a white dress chases persons. & 357 & Contrasting visuals (e.g., baby vs. elder, fire vs. ice) shown side by side in succession.
    \end{tabular}
    \caption{Representative video contents among the 25 best- and worst-predicted cases of emotion estimation.}
    \label{tab:contents_best_worst_of_nishimoto}
\end{table}

\subsubsection*{Similarity structure of all videos}
Next, to evaluate the similarity between the human and the model's similarity structures, we performed the same sets of analyses as we did when comparing the emotion structures of different participant groups in the previous sections. In terms of correlation on each video in the previous section, it was shown that Gemini predicted human emotions with higher accuracy than the level of agreement typically observed among human participants (Figures \ref{fig:nishimoto_human_hist} and \ref{fig:nishimoto_human_Gemini_hist}). However, higher correlations for each video do not necessarily mean that the overall emotion structures of all videos are similar between the model and humans. This section focuses on the structural level comparison between the model and humans based on conventional RSA and GWOT.

\paragraph{Correlation between similarity structures}
We observed a moderately high degree of correlation between representational dissimilarity matrices (RDMs) of all videos in humans and Gemini (Figure \ref{fig:nishimoto_human_Gemini_RDM_GW}A). 
The top panel in Figure \ref{fig:nishimoto_human_Gemini_RDM_GW}A depicts the participants' RDM and the bottom panel depicts Gemini's RDM, both using cosine similarity as metric. Pearson's correlation coefficient between the two RDMs is 0.558, which is significantly higher than the chance level ($[-0.000509\%, 0.00108\%]$, Table \ref{tab:nishimoto_Gemini}), although this value is lower than the correlation coefficient between different participant groups (0.859).

\paragraph{GWOT:One-to-one matching}
Although the correlation between RDMs of humans and Gemini is moderately high (0.558), we found the one-to-one matching rate from GWOT is low, 2.36\% (Table \ref{tab:nishimoto_Gemini}). As we can see in the optimal transportation plan (Figure \ref{fig:nishimoto_human_Gemini_RDM_GW}B), the high values are scattered in places other than the diagonal elements, resulting in the low matching rate. Although this rate is significantly higher than both the theoretical chance level (0.182\%) and the empirically measured chance level ($[1.01\%, 1.23\%]$, Table \ref{tab:nishimoto_Gemini}), it still remains substantially below the matching rate between participant groups (16.36 \%) .

\paragraph{GWOT:category Matching}
Despite the low one-to-one matching rate, we found that category matching rate is reasonably high. 
To analyze category matching, we classified the videos into 10 categories via hierarchical clustering of the participants' emotion reports (see Method for details). From Figure \ref{fig:nishimoto_human_Gemini_RDM_GW}B, we observe that the high values are concentrated within the diagonally aligned boxes outlined by the green lines, indicating a high category matching rate. In Figures \ref{fig:nishimoto_human_Gemini_RDM_GW}A and B, the rows are sorted according to the hierarchical clustering results, and the green lines represent the category boundaries. The model's matching accuracy of all videos based on the optimal transportation plan is 50.5\%, which clearly exceeds the chance level ($[17.3, 19.4\%]$, Table \ref{tab:nishimoto_Gemini}) and is a slightly lower than the category matching rate between participant groups (66.18\%).

\subsubsection*{Similarity structure of selected videos} 
While the results in the previous section indicate that the model fails to capture the emotion structure of all videos adequately, we conducted further analyses on videos with high similarity scores to investigate whether the model might partially capture the structure of some parts of videos. To this end, we selected the top 100 and 250 videos in which the model's predictions showed the highest correlation with human ratings from the results of Figure \ref{fig:nishimoto_human_Gemini_hist}, and conducted the same set of analyses as performed for all videos. Note that the purpose of this additional analysis was not to claim that the GWOT matching rate or RSA correlation increases when the videos are selected, because that is obvious, but rather to use these as a measure to compare performance with others in terms of partial structural alignment.

\begin{figure}
    \centering
    \includegraphics[width = \linewidth]{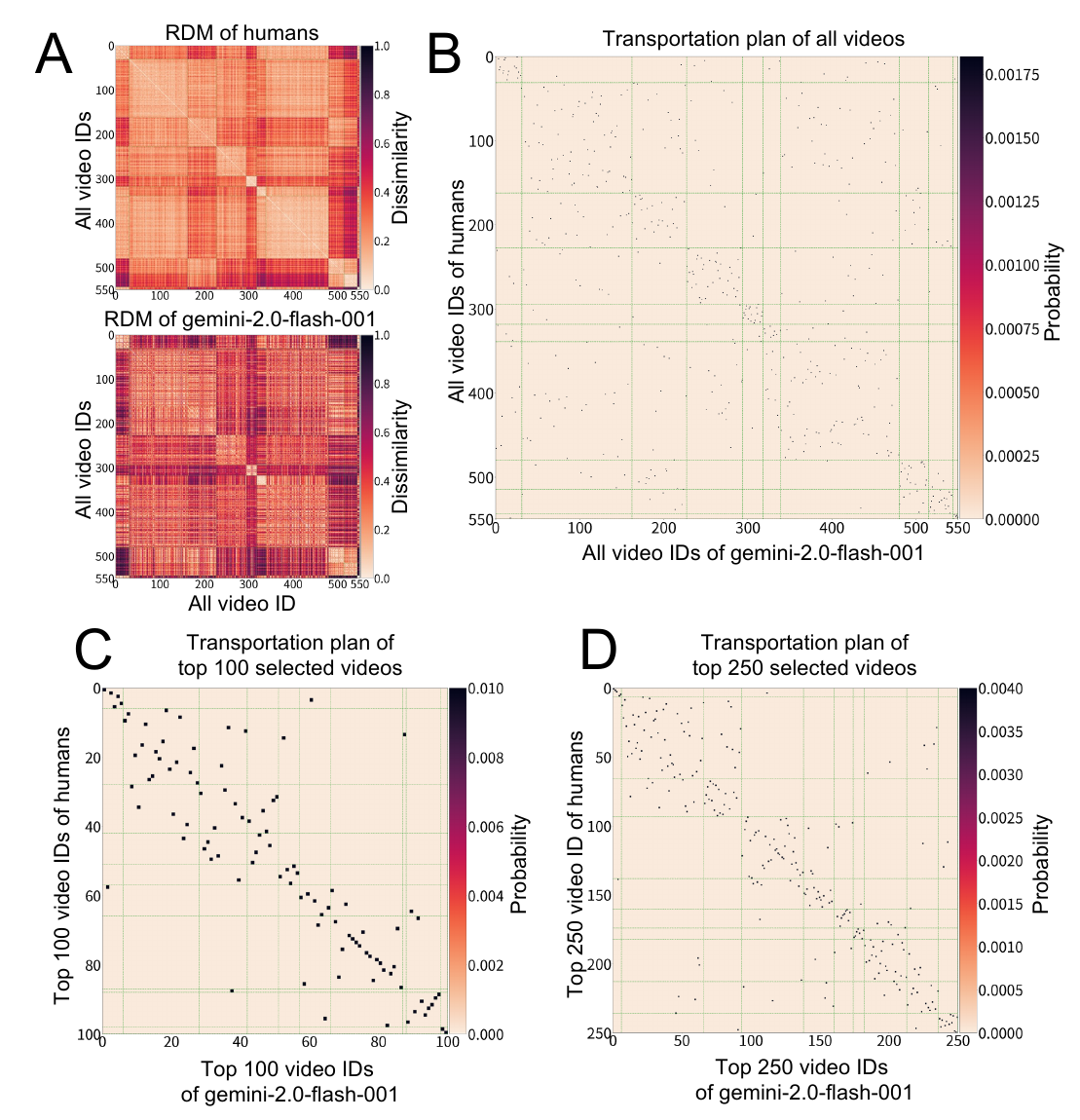}
    \caption{
    Unsupervised comparison of human similarity structures of videos in the Koide-Majima et al. dataset with similarity structures estimated by Gemini based on Gromov-Wasserstein Optimal Transport (GWOT). A: The Representation Dissimilarity Matrices (RDMs) of the human participants and Gemini. The elements of the RDMs represent the dissimilarity between the emotion ratings of the videos, quantified by cosine similarity. B: Optimal transportation plan obtained by GWOT between the human and Gemini RDMs. Green lines represent the category boundaries of the videos. C: Optimal transportation plan for the selected top 100 videos. D: Optimal transportation plan for the selected top 250 videos.
    }
    \label{fig:nishimoto_human_Gemini_RDM_GW}
\end{figure}

Focusing on the top 100 videos and top 250 videos, the model's predicted ratings demonstrated a high degree of structural agreement with human ratings (Figures \ref{fig:nishimoto_human_Gemini_RDM_GW}C, D and Table \ref{tab:nishimoto_Gemini}). Inspection of the transport matrices show that many diagonal elements have non-zero values, indicating a high one-to-one matching rate, and most of the non-zero elements fall within the green, diagonally outlined region, indicating a high category matching rate. To be specific, the one-to-one matching rate was calculated to be 17.0\% for the top 100 videos and 8.4\% for the top 250 videos, respectively, which considerably exceeds the chance level ($[4.90\%, 5.90\%]$ for the top 100 videos and $[2.28\%, 2.60\%]$ for the top 250 videos, Table \ref{tab:nishimoto_Gemini}). The category matching rate reached 69.0\% for the top 100 videos and 71.6\% for the top 250 videos, respectively, likewise far surpassing the chance level ($[19.1\%, 25.5\%]$ for the top 100 videos and $[18.2\%, 21.1\%]$ for the top 250 videos, Table \ref{tab:nishimoto_Gemini}). These results show that, for the selected videos, the emotion structures between humans and models are similar enough that the unsupervised mapping of emotion structure was performed with relatively high accuracy.

\begin{table}[t!]
    \centering
    \begin{tabular}{p{3.0cm}cp{3.0cm}ccc}
        Model&Data pattern&Evaluation method&Top 100 videos&Top 250 videos&All video\\
        \hline
        Human&Video&Mean of correlation on each video&-&-&0.313\\
        \cmidrule(lr){3-6} 
        &&Correlation between RDMs & 0.965 & 0.931& 0.859\\
        \cmidrule(lr){3-6} 
        &&Matching rate of GW alignment &41.0\%&33.6\%& 16.36\%\\
        \cmidrule(lr){3-6} 
        &&10 hierarchical category matching rate&81.0\%&85.6\% &66.18\%\\
        \midrule
        Gemini-2.0-flash-001&Video&Mean of correlation on each video&-&-& 0.374\\
        \cmidrule(lr){3-6} 
        &&Correlation between RDMs & 0.938& 0.818& 0.558\\
        \cmidrule(lr){3-6} 
        &&Matching rate of GW alignment &17.0\%&8.4\%& 2.36\%\\
        \cmidrule(lr){3-6} 
        &&10 hierarchical category matching rate&69.0\%&71.6\% &50.5\%\\
        \midrule
        Shuffled human ratings &Video &Mean of correlation on each video&-&-&$[0.321, 1.72]\times 10^{-2}$\\
        \cmidrule(lr){3-6} 
        &&Correlation between RDMs & $[0.0461, 0.505]$ & $[0.0154, 0.0179]$& $[-0.509, 1.08]\times 10^{-3}$\\
        \cmidrule(lr){3-6} 
        &&Matching rate of GW alignment &$[4.90, 5.90]$\%&$[2.28, 2.60]$\%& $[1.01, 1.23]$\%\\
        \cmidrule(lr){3-6} 
        &&10 hierarchical category matching rate&$[19.1, 25.5]$\%&$[18.2, 21.1]$\% &$[17.3, 19.4]$\%\\
    \end{tabular}
    \caption{Summary of evaluation metrics for each model on the Koide-Majima et al. dataset. For the shuffled human ratings, the 95\% percentile interval of the null distribution is reported for each metric (see Methods for details).
    }
    \label{tab:nishimoto_Gemini}
\end{table}

\subsection*{Evaluation of MLLM's estimation of human emotion structure for data from Cowen\& Keltner (2017)}
To further investigate the MLLMs' performance of emotion ratings, we conducted the same sets of analyses using a different dataset by Cowen \& Keltner \cite{Cowen2017-lx}. In this dataset, 2,185 short video clips (averaging about 5 seconds in length and containing no audio) were presented to participants, who then reported multiple emotions from 34 emotion categories while watching each clip.
As with the Koide-Majima et al. (2020) dataset, we provided the full video input to Gemini-2.0-flash, since it can directly handle entire video sequences. Moreover, because these videos are relatively short, we also adopted an approach where we extracted six frames per video and presented them as input to Molmo-7B-D and GPT-4.1, which do not accept raw video (see Methods for details). This design enabled us to compare how different MLLMs beyond Gemini perform under the same dataset conditions. 
In the following sections, we primarily show results for Gemini and then present the results from the other models for comparison.

By using a similar prompt to that for the Koide-Majima et al. dataset shown in Methods, we obtained the intensity of emotion ratings from Gemini for all video clips in the Cowen \& Keltner (2017) dataset. While Gemini provided responses for all 2,185 videos, one video was excluded from further analysis because all emotion scores were rated as zero, making it impossible to compute cosine similarity with other videos. As a result, 2,184 videos were used for subsequent analyses.
In the following, we quantitatively evaluate similarity between the emotion structure of humans and Gemini, using the conventional correlation RSA and GWOT.

\subsubsection*{Similarity between emotion ratings for individual videos}
By analyzing the correlation on a per-video basis, we found that Gemini's emotion estimates align with participants' ratings at a level exceeding chance (Figure \ref{fig:cowen}A). 
The blue histogram shows the distribution of Pearson correlations between Gemini and human ratings for each video, whereas the gray histogram corresponds to correlations obtained after shuffling the human ratings. The dashed line marks the mean correlation of the original data (0.553), and Cohen's D calculated over 1,000 shuffles was 3.18. Compared to the Koide-Majima et al. dataset, this higher mean correlation and Cohen's D suggest that Gemini estimates video-specific emotions more accurately for this dataset.

\paragraph{Well-Estimated and Poorly-Estimated Videos}
Similarly to the Koide-Majima et al. dataset, we examined the content of the top 25 well estimated videos and those of the bottom 25 badly estimated videos. We selected five representative videos that were particularly typical in content, and summarize them in Table \ref{tab:cowen_contents_best_worst_of_nishimoto}.

Examination of the top 25 videos revealed, consistent with insights from the Koide-Majima et al. dataset, that Gemini exhibits high estimation accuracy for video content that evokes strong emotional responses from a single frame. Specifically, in 1487.mp4 and 1828.mp4, where insects are shown gathering in dirty places, Gemini was able to predict the emotion of ``disgust'' reported by participants. Furthermore, in 0073.mp4, which depicts a needle piercing human skin, and 1362.mp4, which shows a man covered in blood, the model successfully inferred emotions such as ``disgust'' and ``horror.'' These results are consistent with the previous section's conclusion that videos whose emotions can be clearly triggered by visual information alone tend to yield higher estimation accuracy.

In contrast, an analysis of the bottom 25 videos with the lowest correlation revealed two main reasons why Gemini may have failed to accurately estimate emotions—similar to the findings with the Koide-Majima et al. dataset.

First, Gemini tended to struggle with videos that required contextual understanding or background reasoning. For instance, in 2130.mp4, a person narrowly avoids being hit by a car while sledding. While participants predominantly reported feelings of ``relief,'' the model's strongest prediction was ``amusement.'' This discrepancy shows that the model had difficulty capturing the contextual nuance of a near-miss accident. Additionally, in 0882.mp4, a player attempts to high-five a teammate but is ignored. While participants commonly reported the emotion ``awkward,'' the model predicted emotions such as ``interest'' and ``admiration.'' This shows that the model struggled to interpret the subtle social context of the discomfort or embarrassment resulting from a rejected social interaction.

Second, some videos make it difficult even for human viewers to pinpoint a single, clear emotion. For example, 0664.mp4 (featuring birds hung upside down) and 0888.mp4 (depicting writhing tentacles) induce a vague sense of discomfort, yet participants themselves may be uncertain which emotion to report. Such ambiguity can lead to substantial variability in human ratings, thereby complicating the model's ability to accurately estimate emotions.

\begin{table}
    \centering
    \begin{tabular}{c|p{6.4cm}||c|p{6.4cm}}
        \multicolumn{2}{c||}{Well estimated videos} & \multicolumn{2}{c}{Badly estimated videos} \\ \hline
        Video ID & Contents & Video ID & Contents \\ \hline
        0073 & A needle pierces the skin.
        & 0882 & Tries to high-five a teammate but is ignored.\\ 
        1362 &A blood-covered man is sitting on the ground.
        & 0888 & Tentacles are being grilled on a net\\
        1487 & Multiple hidden insects suddenly appear in the bathroom.
        & 1916 & Throwing a brick into the middle of a swarm of insects causes them to scatter.\\ 
        1689& A bird is flying through the gaps between trees.
        & 2083& A police officer, noticing he's in a selfie, exits the frame with a bizarre move.\\
        1828 & The camera zooms in from a swarm of insects to focus on a single insect.
        & 2130 & A person sledding down a snowy mountain narrowly avoids a colliding with a car.
    \end{tabular}
    \caption{Contents of the five best-estimated videos and the five worst-estimated videos from data of Cowen \& Keltner(2017).}
    \label{tab:cowen_contents_best_worst_of_nishimoto}
\end{table}

\subsubsection*{Similarity structure of all and selected videos}
\begin{figure}
    \centering
    \includegraphics[width=\linewidth]{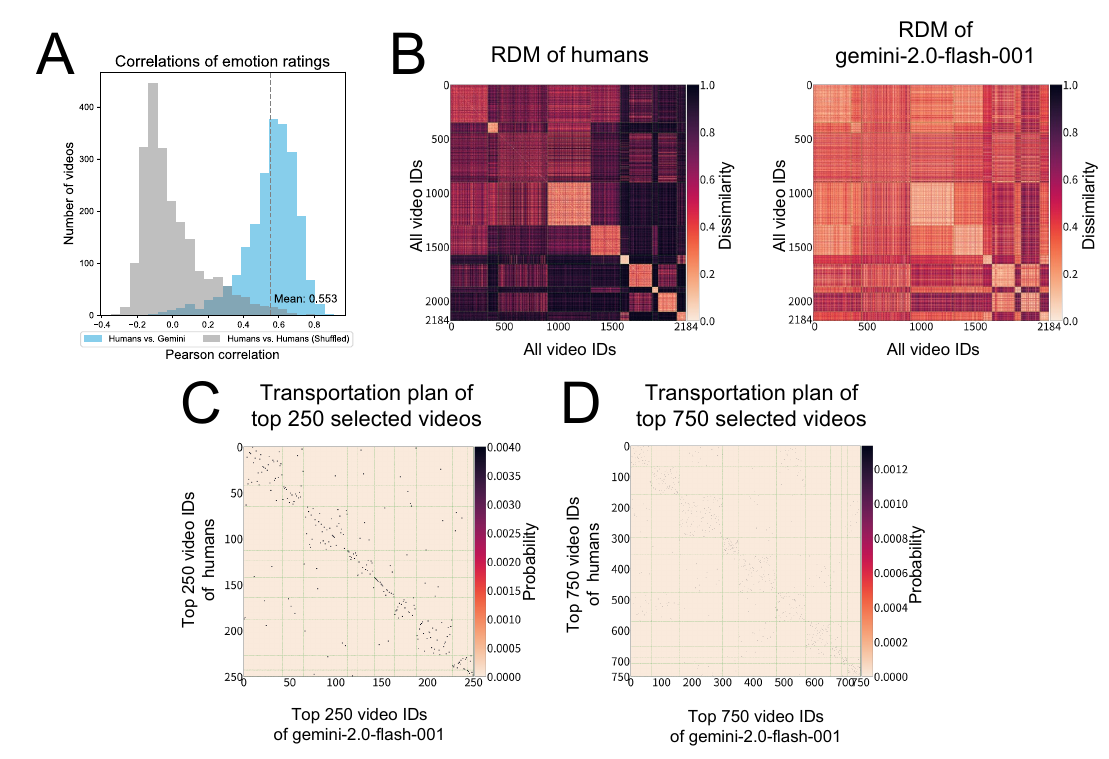}
    \caption{Comparison of the human similarity structures of videos in the Cowen \& Keltner dataset with the similarity structures estimated by Gemini. 
    A: Histograms of the Pearson correlation of each video clip between the human ratings and Gemini's estimation. The blue histogram represents the distribution of correlation between the human ratings and Gemini's estimation for each video, and the gray histogram represents the distribution of correlation between the human ratings and the shuffled human ratings, which served as the null distribution. The dashed line represents the mean of the correlation, 0.553, between the human ratings and Gemini's estimation (blue histogram).
    B: Representation Dissimilarity Matrices (RDMs) of the human participants and Gemini. The elements of the RDMs represent the dissimilarity between the emotion ratings of the videos, quantified by cosine similarity.
    C: Optimal transportation plan obtained for the selected top 250 videos by GWOT between the human and Gemini RDMs. Green lines represent the category boundaries of the videos. 
    D: Optimal transportation plan for the selected top 250 videos.
    }
    \label{fig:cowen}
\end{figure}

To evaluate the similarity structure between participants and Gemini, we conducted two analyses, RSA and GW alignment (Figure \ref{fig:cowen}B, C, D, Table \ref{tab:CK_table}).

\paragraph{Correlation between similarity structures}
We observed a moderately high degree of correspondence between the representational dissimilarity matrices (RDMs) of humans and Gemini for all videos in this dataset (Figure \ref{fig:cowen}B). The left panel in Figure \ref{fig:nishimoto_human_Gemini_RDM_GW} shows the participants' RDM and the right panel shows the RDM of Gemini; both use cosine similarity as their metric. Notably, this dataset is approximately four times larger than the Koide-Majima et al. dataset, yet we still obtained a correlation coefficient of 0.555 — a reasonably high value for such a sizable collection of videos. The Pearson correlation of 0.555 is well above chance (Table \ref{tab:CK_table}), showing that Gemini can approximate the overall emotion structure at an even larger scale than that examined in the Koide-Majima et al. dataset.

\paragraph{GWOT: One-to-one matching}
Despite the high correlation of RDMs between humans and Gemini (0.558), we found that the one-to-one matching rate was low, at 1.69 \% (Figure S1, Table \ref{tab:CK_table}), which is close to the chance level (0.229\%).

We then also evaluated the similarity structures of selected videos, using the top 250 and 750 videos by Pearson correlation to investigate the possibility that Gemini accurately captures the similarity structures for selected videos (Figures \ref{fig:cowen}C, D). Both panels show that there are many non-zero values at the diagonal elements and the matching rate is 18.8 \% for the top 250 videos and 7.47 \% for the top 750 videos, which are significantly higher than the chance level ($[3.01\%, 3.46\%]$ and 0.667\%, respectively).

This result shows that Gemini did not estimate the overall structure of the videos well, but was able to estimate the structure of a part of the videos well enough that unsupervised mapping is possible to some extent. This result is consistent with the results of the Koide-Majima et al. dataset and strengthens the robustness of the result that Gemini is able to estimate the high-dimensional structure of human emotions.

\paragraph{GWOT: category matching}
Despite of the low one-to-one matching rate for all videos, we found that the category matching rate was reasonably high, as similarly observed in analysis of the Koide-Majima et al. dataset. For the analysis of category matching, we classified the videos into 10 categories via hierarchical clustering of the participants' emotion reports (see Methods for details). In Figure \ref{fig:cowen}B, C, D and Figure S1, the rows are sorted according to the hierarchical clustering results, and the green lines indicate category boundaries.
As shown in Figures S1, \ref{fig:cowen}C and D, the matching values are concentrated along the diagonally outlined boxes (green lines), indicating a high level of category alignment. The category matching rates were 78.8\% for the top 250 videos and 69.2\% for the top 750 videos, both substantially above the chance level ($[19.1\%, 25.4\%]$ and 21.2\%, respectively; see Table \ref{tab:CK_table}).
The category matching rate for all videos was 54.6\%, which also significantly exceeds the chance level (15.6\%; see Table \ref{tab:CK_table}).
These findings demonstrate that Gemini broadly captured the emotion structure of all the videos at the coarse-category level when watching the videos.

\subsubsection*{Comparison with other MLLMs}
In this section, we conducted the same experiment using GPT-4.1\cite{Open2025-fy} and Molmo-7B-D\cite{Deitke2024-tz}, two of the latest MLLMs, to evaluate their emotion estimation accuracy. Since these models cannot process video input directly, we adopted an approach where multiple frames were extracted from each video and input as still images to enable a fair comparison (See Methods, `Selection model for data from Cowen \& Keltner (2017)', for details).
As a result, we obtained responses for 2,184 videos from GPT and 1,973 videos from Molmo, and used these data for further analysis.

GPT-4.1 achieved a performance roughly comparable to or slightly below that of Gemini (Table \ref{tab:CK_table}). For instance, its RSA value was 0.486 — about 0.06 lower than Gemini's. When focusing on the top 250 or top 750 videos selected based on the correlation between each model's predictions and human ratings, GPT-4.1 achieved one-to-one matching rates of 16.8\% for the top 250 and 4.67\% for the top 750 videos, and category-level matching rates of 68.6\% for the top 250 and 57.6\% for the top 750 videos. While these values are slightly lower than those of Gemini, they can still be considered very favorable results.
These observations indicate that although GPT-4.1 does not achieve the same level of accuracy as Gemini, it appears to capture human emotion structure to a reasonable degree in one-to-one matching and to quite a high degree in category matching.

Meanwhile, although Molmo-7B-D exceeded the chance level on all evaluation metrics, its accuracy remained lower than that of Gemini and GPT-4.1 (Table \ref{tab:CK_table}). For example, although its RSA value was 0.229 — over 0.2 points below Gemini's — it was still clearly above the chance level. Furthermore, in analyses focusing on the top 250 and top 750 videos selected based on the correlation between each model's predictions and human ratings, Molmo-7B-D's performance in category matching was 36.8\% for the top 250 videos and 36.9\% for the top 750 videos. Although these values exceed the chance level ($[19.1\%, 25.4\%]$ and 21.2\%, respectively), they remain lower than those of Gemini and GPT-4.1.

\begin{table}[t!]
    \centering
    \begin{tabular}{p{3.0cm}cp{3.0cm}ccc}
        Model&Data pattern&Evaluation method&Top 250 videos&Top 750 videos&All video\\
        \hline
        Gemini-2.0-flash-001&Video&Mean of correlation on each video&-&-&0.553\\
        \cmidrule(lr){3-6} 
        &&Correlation between RDMs & 0.836 & 0.771& 0.555\\
        \cmidrule(lr){3-6} 
        &&Matching rate of GW alignment &18.8\%&7.47\%& 1.69\% \\
        \cmidrule(lr){3-6} 
        &&10 hierarchical category matching rate &78.8\%& 69.2 \%& 54.6\% \\
        \midrule
        GPT-4.1-2025-04-14&6 frames&Mean of correlation on each video&-&-&0.538\\
        \cmidrule(lr){3-6} 
        &&Correlation between RDMs & 0.818& 0.714& 0.486\\
        \cmidrule(lr){3-6} 
        &&Matching rate of GW alignment & 16.8\%& 4.67\%&  0.795\%\\
        \cmidrule(lr){3-6} 
        &&10 hierarchical category matching rate & 68.8\%& 57.6\%&  45.8\%\\
        \midrule
        Molmo-7B-D-0924&6 frames&Mean of correlation on each video&-&-&0.21\\
        \cmidrule(lr){3-6} 
        &&Correlation between RDMs & 0.480 & 0.430& 0.229 \\
        \cmidrule(lr){3-6} 
        &&Matching rate of GW alignment &4.40\%&1.60\%& 0.304\%\\
        \cmidrule(lr){3-6} 
        &&10 hierarchical category matching rate &36.8\%&36.9\%& 20.3\%\\
        \midrule
        Shuffled human ratings&Video&Mean of correlation on each video&-&-&$[-0.930, 1.26]\times10^{-2}$\\
        \cmidrule(lr){3-6} 
        &&Correlation between RDMs & $[0.201, 0.203]$ & $[0.0533, 0.0543]$ & $[-1.09, 2.43]\times 10^{-4}$ \\
        \cmidrule(lr){3-6} 
        &&Matching rate of GW alignment &$[3.01, 3.46]$\%&0.667\%& 0.229\%\\
        \cmidrule(lr){3-6} 
        &&10 hierarchical category matching rate &$[19.1, 25.4]$\%&21.2\%& 15.6\%
    \end{tabular}
    \caption{Summary of evaluation metrics for each model on the Cowen \& Keltner dataset. 
    For the shuffled human ratings, the 95\% percentile interval of the null distribution is reported for each metric (see Methods for details). Those without 95\% intervals are based on a single shuffle due to computational constraints.
}
    \label{tab:CK_table}
\end{table}

\section*{Discussion}
This study investigated the extent to which MLLMs can reproduce the high-dimensional structure of human emotions. In a supervised approach using RSA, Gemini and GPT showed high consistency with human emotional structures and good performance. The open-weight model Molmo showed lower performance than these models, but still achieved a reasonable level of reproduction. In the unsupervised approach using GWOT, while strict one-to-one matching over the entire set of videos remained challenging, the models were able to achieve a reasonably high level of matching at the category level. Taken together, these results indicate that modern MLLMs are capable of capturing substantial portions of human emotion structure, but have not yet reached the level that would allow for proper one-to-one matching at the fine item level. This mixed profile sets the context for the remainder of the Discussion, where we explore how recent advances – particularly in GPT and Gemini – have solidified the core perceptual capabilities of accurately recognizing what is depicted in the video, ``Step 1,'' and thereby enabled progress toward the more challenging goal of inferring latent emotion structures, ``Step 2,'' based on the inference of ``Step 1.'' We discuss the cases in which current models accurately estimate human emotional responses, and those in which current models do so relatively poorly.

Based on the results of this study, along with benchmarking results on various tasks from early models to the latest MLLMs, we first emphasize that the rapid and significant technological leap in overall MLLM performance from the initial models that appeared around 2023 to the state-of-the-art models of 2025 has for the first time allowed the level of accuracy in emotion inference demonstrated in this study. Early MLLMs such as BLIP2\cite{Li2023-ex} and LLaVA-1.5\cite{Liu2023-qz}, which appeared around 2023, demonstrated the potential of multimodal processing, but faced numerous challenges in terms of visual recognition accuracy, instruction following, and contextual understanding\cite{noauthor_undated-kp}. The subsequent release of GPT-4V marked significant progress in benchmarks \cite{OpenAI2023-mo, Lian2024-fb}. In our evaluations, we began to observe responses that highly correlated with human emotion ratings. However, GPT-4V still exhibited inaccuracies in visual recognition and contextual understanding, often misclassifying innocuous content as inappropriate (e.g., mislabeling it as sexual or violent) \cite{Wang2025-jn}. Due to misclassification, the percentage of videos for which GPT4-V outputs a responses is much lower (507/2185, 23\%) than that of GPT-4.1 (2184/2185, 99\%). For this reason, we judged that a comparison with GPT4-V would be difficult and decided not to include GPT-4V in our analysis. With the introduction of GPT-4.1 and Gemini-2.0-flash in 2025, the accuracy of Step 1 improved dramatically, as indicated in benchmarking results \cite{Open2025-fy, Mallick2025-ba}. This advance has enabled accurate visual recognition and effective handling of complex instructions, and provided a solid foundation for emotion inference. As a result, the challenging and ambiguous task of inferring emotional structure (Step 2) became practically feasible. In our evaluations, misclassifications were significantly reduced and consistent responses were obtained across many videos. Notably, we were able to achieve stable, high correlations with human emotional structures, as demonstrated in this study. These results highlight that MLLMs have undergone rapid evolution in both Step 1 (recognition and comprehension) and Step 2 (emotion inference), signaling not just incremental improvements but clear transition to a stage where multimodal AI can address complex and context-dependent tasks.

At the same time, this study also revealed that even models with excellent recognition capabilities (Step 1) still face limitations when handling the integrated challenge of emotion structure inference (Step 2), indicating both the difficulty of the task and the exceptional capabilities of state-of-the-art models.
For example, Molmo-7B-D, which we used as a comparison model, demonstrated outstanding performance in Step 1 across multiple benchmarks \cite{Deitke2024-tz}. Molmo-7B-D outperformed GPT-4V in the multimodal benchmark\cite{Deitke2024-tz} and even surpassed then-state-of-the-art models like Gemini 1.5 Pro and GPT-4o-0513 in tasks such as VAQ v2 \cite{Goyal2017-yv} and TextVQA \cite{Singh2019-ed}. However, emotion structure inference (Step 2) cannot be solved by the simple application of recognized information. It requires estimating latent emotional structures that are inherently subjective and context-dependent. Therefore, high accuracy in Step 1 alone is insufficient; models must also be able to interpret recognized information within deep contextual frameworks to derive emotional meaning. While Molmo exhibited strong performance in Step 1, it showed a clear performance gap compared to GPT and Gemini when tackling tasks involving latent emotion inference. This does not indicate a deficiency in Molmo's capabilities but rather highlights that such tasks demand more than recognition accuracy or general reasoning, but rather complex, multi-layered processing.
Notably, GPT and Gemini achieved further improvements and stability in Step 1, enabling them to effectively manage this previously challenging integrated task (Step 2). The fact that these SOTA models successfully perform tasks where even highly capable models like Molmo struggle strongly suggests that MLLMs have advanced beyond gradual improvements, reaching a new stage where AI can address integrated and context-sensitive challenges.

To better understand the nature of these improvements, we examined cases in which emotion estimation worked particularly well. These analyses revealed that MLLMs exhibit high predictive accuracy from a structural perspective, particularly for emotions that do not strongly depend on contextual cues. For instance, videos featuring babies or cats and dogs consistently elicited predictions of ``cuteness'' or ``fondness.'' Similarly, scenes showing a long-haired woman in a white dress in a dimly lit setting repeatedly triggered predictions of ``fear.'' These results suggest that MLLMs are capable of appropriately reconstructing emotional responses primarily driven by visual features \cite{Kragel2019-is}.

On the other hand, MLLMs struggled to estimate emotions in videos that relied heavily on contextual cues. For example, in one video, a player attempts a high-five with a teammate and is ignored, while other players in the background are seen successfully high-fiving. Visually, this scene might suggest a positive emotion such as ``admiration.'' However, when considering the broader context, a more accurate emotional interpretation would be ``awkwardness.'' These findings indicate that emotional understanding cannot be derived from visual features alone; it requires the interpretation of situational and social context. When MLLMs fail to appropriately interpret these contextual cues, their emotion predictions tend to diverge from human judgments.

In addition to contextual cues, it would be beneficial to design models that can integrate not only social contextual information but also interoceptive signals such as heart rate and bodily sensations, which are known to essentially affect emotion perceptions \cite{Barrett2017-vn,Barrett2017-ex, Ohira2020-jv}. 
Our findings indicate that while current MLLMs perform well in estimating emotions based on explicit visual features, they still face limitations in understanding more complex forms of context, such as social relationships and temporal dynamics. Crucially, such contextual understanding involves not only external cues but also internal bodily states. In fact, some of the videos that were particularly challenging for the models involved emotions, such as ``heart-pounding.'' These types of emotions might not be fully interpreted through visual input alone and require sensitivity to interoceptive signals in combination with environmental and social factors. Therefore, future advances in MLLMs should involve incorporating diverse modalities into the training process to enable more human-like, context-sensitive emotion understanding.

However, another possibility is that the relatively lower accuracy observed for bodily-driven emotions could partly stem from insufficient representation in pretrained datasets rather than solely the absence of interoceptive signal integration. Emotions strongly influenced by bodily sensations may be less frequently articulated explicitly in textual form, possibly limiting their prevalence in standard pretrained data. Consequently, incorporating direct interoceptive signals into the model architecture might not be strictly necessary. Instead, it could be beneficial to fine-tune existing pretrained models on new datasets specifically designed to better represent these emotion-scene associations. Considering the promising performance of MLLMs demonstrated by our zero-shot results, fine-tuning might provide a viable path forward. Future research should investigate whether fine-tuning pretrained models can lead to improvements in prediction accuracy and unsupervised alignment for these potentially challenging emotional contexts.

Taken as a whole, the fact that MLLMs are beginning to reproduce the structural patterns of emotions induced by visual features represents a significant advance in affective understanding. However, challenges remain in estimating emotions that are highly dependent on contextual and interoceptive factors, and full replication of human emotional experience has yet to be achieved. Nevertheless, the emerging ability of MLLMs to approximate emotional structures that were previously difficult to model suggests a new frontier for affective computing and offers a crucial foundation for future research and practical applications.

\section*{Methods}
\subsection*{Emotion ratings during video viewing}
We analyzed two datasets of emotion ratings during video viewing from previous studies. One is from Koide-Majima et al. (2020) and the second is from Cowen and Keltner (2017). The details of each experiment and the data obtained should be referred to in the original papers, but here we provide brief information necessary to understand the present paper.

\subsubsection*{Data from Koide-Majima et al. (2020)}
\paragraph{Videos} In total, 550 different video clips were presented, chosen to evoke a diverse range of emotional responses. The genres of the selected videos included horror, violent drama, comedy, romance, fantasy, everyday life scenes, and action. The duration of each video clip was 10–20 seconds, and about 15 seconds on average.

\paragraph{80 emotion categories used for subjective reports}
80 emotion categories were used from various sources to cover a wide range of emotions. Note that these emotion categories used Japanese words for Japanese participants in the experiments. The English translations of the 80 Japanese emotion categories are as follows: (1) love, (2) amusement, (3) craving, (4) joy, (5) nostalgia, (6) boredom, (7) calmness, (8) relief, (9) romance, (10) sadness, (11) admiration, (12) aesthetic appreciation, (13) awe, (14) confusion, (15) entrancement, (16) interest, 
(17) satisfaction, (18) excitement, (19) sexual desire, (20) surprise, (21) nervousness, (22) tension, (23) anger, (24) anxiety, (25) awkwardness, (26) disgust, (27) empathic pain, (28) fear, (29) horror (bloodcurdling), (30) laughing, (31) happiness, (32) friendliness, (33) ridiculousness, (34) affection, (35) liking, (36) shedding tears, (37) emotional hurt,(38) sympathy, (39) lethargy, (40) empathy, (41) compassion, (42) curiousness, (43) unrest, (44) exuberance, (45) appreciation of beauty, (46) fever, (47) scare (feel a chill), (48) daze, (49) positive-expectation, (50) throb, (51) sexiness, (52) indecency, (53) embarrassment, (54) oddness, (55) contempt, (56) alertness, (57) eeriness, (58) positive-emotion, (59) vigor, (60) longing, (61) tenderness, (62) pensiveness, (63) melancholy, (64) relaxedness, (65) acceptance, (66) unease, (67) negative-emotion, (68) hostility, (69) levity, (70) protectiveness, (71) elation, (72) coolness, (73) cuteness, (74) attachment, (75) encouragement, (76) annoyance, (77) positive-fear, (78) aggressiveness, (79) distress, and (80) stress.

\paragraph{Emotion ratings from participants}
166 Japanese annotators rated emotions while viewing 550 video clips. Each annotator was instructed to rate how well an emotion category (e.g., ``laughing'') matched their personal feelings elicited by the video scene, using a scale from 0 (not matched at all) to 100 (perfectly matched). Importantly, they were told to base their ratings on their own emotional responses, not those of the characters in the videos. During the rating process, annotators continuously indicated the degree of matching by moving a mouse while viewing the video stimuli, with ratings recorded at one-second intervals. For each of the 80 emotion categories, they obtained four independent ratings by assigning four different annotators to each category. To accomplish this, a total of 166 annotators participated, with each person rating one or two emotion categories. When an annotator was assigned two categories, they first rated one emotion (e.g., ``disgust'') throughout the entire set of video clips. They then watched the entire sequence again to rate the second emotion category (e.g., ``satisfaction'').

\subsubsection*{Data from Cowen \& Keltner (2017)}
\paragraph{Videos}
These investigators collected 2{,}185 short videos to cover a wide range of emotion-elicited situations. 
On average, each clip was about 5 seconds long.

\paragraph{34 emotion categories}
34 English emotion categories were selected from emotion taxonomies of prominent theories and used in the experiments. The 34 emotion categories were: (1) admiration, (2) adoration, (3) aesthetic appreciation, (4) amusement, (5) anger, (6) anxiety, (7) awe, (8) awkwardness, (9) boredom, (10) calmness, (11) confusion, (12) contempt, (13) craving, (14) disappointment, (15) disgust, (16) empathic pain, (17) entrancement, (18) envy, (19) excitement, (20) fear, (21) guilt, (22) horror, (23) interest, (24) joy, (25) nostalgia, (26) pride, (27) relief, (28) romance, (29) sadness, (30) satisfaction, (31) sexual desire, (32) surprise, (33) sympathy, and (34) triumph.

\paragraph{Emotion ratings from participants}
The participants selected multiple emotions from the 34 English emotion categories elicited by each video. Participants were instructed to choose at least one emotion category but could choose as many as desired. Each of the 2,185 videos was judged by 9 to 17 observers. The ratings of each emotion category were averaged for each video.

\subsection*{Dealing with human ratings}
\subsubsection*{Data from Koide-Majima et al. (2020)}
\paragraph{Splitting participants into two groups}
To investigate the extent to which emotion structures are shared among humans, we split the dataset into two groups. For each emotion category, four participants had provided ratings; these were divided into two pairs, yielding two independent participant groups per category. Within each group, we then averaged the two participants' ratings for the same emotion category, treating each group as an independent observer. 

Although we wanted to make the ratings of the two groups as independent as possible, the limitations of the experimental data make it impossible to separate the two groups so that there is no overlap of participants. Note, however, that within any single emotion category, there is no overlap of participants between the two split groups. Thus, although the split used in this study does not perfectly guarantee independence in terms of overlap of participants, which would increase the similarity between two groups, there is some degree of independence for ratings at the level of each emotion category.

\paragraph{Averaging the emotion ratings}
First, we averaged the reported ratings for each emotion category. Specifically, if the data were split into two groups, the average was taken from two participants' ratings per category; if the data were not split, four participants' ratings were averaged. Next, based on these averaged values, we further aggregated the second-by-second emotion ratings on a per-video basis. Consequently, each video ended up with 80 averaged emotion ratings, which were then used for the analysis.

\subsubsection*{Data from Cowen \& Keltner (2017)}
Cowen's published emotion ratings represent the proportion of evaluators who selected each emotion category for each video. In this study, we treated this proportion as the intensity of the emotion experienced by participants and performed our analyses accordingly. Moreover, only the average frequency of emotions across all participants was publicly available, and no participant-level data were disclosed. This limitation prevented any comparison of emotion structures among individual participants. Consequently, for this experiment, we performed analyses solely between humans and the model.

\subsection*{Selection of Multimodal LLMs}
Based on the results of preliminary evaluations, we selected Gemini-2.0-flash, GPT-4.1, and Molmo-7B-D as the multimodal Large Language Models (MLLMs) for use in this study. Gemini and GPT 
were chosen because they consistently rank at the top in multiple benchmark tests \cite{Puatruaucean2023-uu, xAI2024-ba} and the latest Chatbot Arena leaderboard as of April 2025 \cite{Chiang2024-ta}, demonstrating high accuracy in predictions for multimodal inputs, including videos and images. However, as commercial models, their internal architectures and training data remain undisclosed. In contrast, to evaluate the performance of open-source MLLMs, we selected Molmo-7B-D\cite{Deitke2024-tz}, which has been reported to achieve performance comparable to Gemini-1.5-Pro and GPT-4o across several benchmarks (Table 1 in \cite{Deitke2024-tz}). In particular, we found that Molmo exhibited strong instruction-following capabilities in our task. Additionally, due to computational resource constraints, we adopted the lightweight 7B configuration instead of larger models such as the 70B variant. Although Llama 3\cite{Grattafiori2024-sh} is also recognized as an excellent open-source model, it was excluded from this study to limit the computational cost associated with GWOT analysis, which increases with the number of models evaluated. Although the above models were the target models in this study, due to data set limitations (explained in the following section), we used only Gemini for the Koide-Majima et al. dataset and Gemini, GPT-4.1, Molmo-7B-D for the Cowen \& Keltner dataset.

\subsubsection*{Selected model for data from Koide-Majima et al. (2020)}
We used \texttt{Gemini-2.0-flash-001} for analysis of the Koide-Majima et al. dataset.
Because the Koide-Majima et al. dataset consists of videos averaging 15 seconds in length that include audio and were rated by Japanese participants, the model used must be able to process videos containing audio and produce output in Japanese. Among the three models discussed in the previous section, only \texttt{Gemini-2.0-flash-001} meets all of these criteria. Consequently, we employed this model only for our analyses of the Koide-Majima et al. dataset.

\subsubsection*{Selected model for data from Cowen \& Keltner (2017)}
The Cowen \& Keltner dataset comprises short, audio-free videos of about five seconds in length, rated by English-speaking participants. For this study, we selected \texttt{Gemini-2.0-flash-001}, \texttt{GPT-4.1-2025-04-14}, and \texttt{Molmo-7B-D-0924} as suitable models for processing this dataset.
Given that these videos have no audio and are only about five seconds in length, we deemed it feasible to capture nearly the same amount of information by extracting multiple frames and assembling them into a single image. Therefore, any model capable of image input and handling English was considered appropriate. On this basis, we adopted GPT-4.1 and Molmo-7B-D. Meanwhile, although \texttt{Gemini-2.0-flash-001} can also handle videos with audio, it can process this dataset without issue, as the Cowen \& Keltner data involve short, audio-free videos rated by English speakers. Consequently, we compared the performances of the three models—Gemini-2.0-flash-001, GPT-4.1-2025-04-14, and Molmo-7B-D-0924 —on the Cowen \& Keltner dataset.

\paragraph{Inputting the videos into Molmo and GPT}
Because Molmo cannot accept direct video input, we extracted six frames from each video and concatenated them horizontally into a single image for its input. By contrast, GPT can handle multiple images simultaneously, so we provided all six frames at once without merging them into a single image.
Specifically, the video was divided into six equal segments, and the first frame of each segment was selected. These six frames were then concatenated horizontally to form one continuous image input for Molmo.

The choice of using six frames was determined after exploring how many frames could be placed within a single image such that each frame remained clearly identifiable by the model. Using too few frames might omit critical information from the video, whereas including too many frames could reduce image resolution or increase computational overhead. Ultimately, six frames provided a suitable balance between retaining essential information and managing resource constraints.

\subsection*{Collecting responses from Models}
We collected responses (outputs) from the models (Gemini-2.0-flash, GPT-4.1, Molmo-7B-D) to evaluate each output using the following procedure. The details of these procedures are described in the following sections. All models were used with their default parameter settings provided by the respective APIs or repositories, without any additional fine-tuning or modification.

\paragraph{Multiple Response Retrieval} For non-sensitive videos, we used the same prompt up to three times and took the average of the three outputs obtained from the model. A preliminary experiment indicated that adding a fourth or additional responses did not improve the correlation with human ratings; hence, we limited retrieval to three outputs to maintain efficiency.

\paragraph{Handling Sensitive Videos} 
Some videos contained sensitive content (e.g., violence, sexual themes), and in such cases, the model might not provide any response. However, there was a probabilistic chance of obtaining a valid output. Therefore, for videos deemed sensitive, we input the same prompt up to ten times and, if at least one valid response was generated, we adopted that response for our analysis. If no response was acquired after ten attempts, the video was excluded from the dataset.
This approach aimed to reduce variability in model outputs and maximize the likelihood of obtaining responses even for sensitive content.

\subsubsection*{Data from Koide-Majima et al. (2020)}
To obtain emotion ratings from the model, we carefully aligned both the input and output formats with the experimental conditions used for human participants. For visual input, each video in the Nishimoto dataset was directly presented to the model to generate emotion intensity predictions. The model used, Gemini, supports video input; therefore, no preprocessing such as frame extraction or conversion to still images was necessary, and the original video files were used as-is. Additionally, the scale for emotional intensity ratings was matched to that used by human participants: the model was instructed to output a score ranging from 0 (not matched at all) to 100 (perfectly matched) for each emotion category. This approach aimed to reduce variability in model outputs and maximize the likelihood of obtaining a response for even sensitive content.

To obtain emotion ratings from the model, we tested multiple prompt formats. We explored variations such as presenting all emotion categories at once or splitting them into multiple prompts, as well as including role-inducing expressions. Within the range we tested, these differences did not lead to substantial variation in the model's output performance. Therefore, we adopted a prompt that was as simple as possible and closely aligned with the instructions used in the actual human experiment. Additionally, since the participants' evaluations were conducted in Japanese, the prompt was also presented to the model in Japanese. Specifically, we translated the following English instruction into Japanese for use with the model:

\begin{tcolorbox}[colframe=black, colback=white, sharp corners=southwest, boxrule=1pt, width=\textwidth, title = Prompt]

\textbf{Please watch the video clip provided.} 
This is a short video clip. Please estimate the intensity of each emotion category listed below that people might feel when watching this video, according to the rating rules given below.

\bigskip
\textbf{Emotion Categories:} \\
love, amusement, craving, joy, nostalgia, boredom, calmness, relief, romance, sadness, admiration, 
aesthetic appreciation, awe, confusion, entrancement, interest, satisfaction, excitement, sexual desire, surprise,
nervousness, tension, anger, anxiety, awkwardness, disgust, empathic pain, fear, horror, laughing, happiness, 
friendliness, ridiculousness, affection, liking, shedding tears, emotional hurt, sympathy, lethargy, empathy,
compassion, curiousness, unrest, exuberance, appreciation of beauty, fever, scare, daze, positive-expectation, 
throb, sexiness, indecency, embarrassment, oddness, contempt, alertness, eeriness, positive-emotion, vigor, longing,
tenderness, pensiveness, melancholy, relaxedness, acceptance, unease, negative-emotion, hostility, levity, 
protectiveness, elation, coolness, cuteness, attachment, encouragement, annoyance, positive-fear, aggressiveness, distress, stress.

\bigskip

\textbf{Rating Rules:}
\begin{itemize}[noitemsep, topsep=0pt]
    \item Rate the intensity of each emotion that people might feel upon both the images and the sounds that make up the scene of the video.
    \item Rate the intensity of each emotion on a scale of 0 to 100, where 0 indicates \textit{`not matched at all'} and 100 indicates \textit{`perfectly matched'}.
    \item Pay attention to the trivial connections between each emotion and the scene of the video, and rate them as carefully as possible.
\end{itemize}

Please rate each emotion individually, following this format:

\texttt{love: [numerical value 0-100] \\
amusement: [numerical value 0-100] \\
... \\
stress: [numerical value 0-100]
}

Respond with numerical values only for each emotion, without additional explanation.
\end{tcolorbox}
\subsubsection*{Data from Cowen \& Keltner (2017)}
In this study, we obtained emotion intensity predictions by inputting each video along with a prompt into the model. For models that supported video input, the original video files were used directly as visual input without any preprocessing. For models that did not support video input, we instead used still images created by extracting frames from the videos (see Methods, ``Selection of Multimodal LLMs'', for details.)

Additionally, based on the emotion rating format of the Cowen \& Keltner dataset, we designed the model's output format to be comparable to the aggregated human evaluations. The Cowen \& Keltner dataset does not provide individual-level participant ratings; instead, it only offers the proportion of raters who selected each emotion category for each video. Strictly replicating this format would require collecting multiple responses from the model for each video, which is impractical in terms of both cost and labor. Therefore, in this study, we interpreted these proportions as emotion intensity rating and instructed the model to output a single intensity score for each emotion category (see Methods, `Dealing with human ratings', for details).

We also explored several variations in prompt format. Similar to the Koide-Majima et al. dataset, we compared presenting emotion categories all at once versus separately, and whether to include role-defining expressions.
Since the intensity scale for emotions can be arbitrarily set, we tried several options, but found no substantial differences in output. Therefore, we adopted a 0–9 scale in this study and normalized the resulting scores to a 0–1 range by dividing them by 10.

Given that the original experiment with human participants was conducted in English, we inputted the following prompt:
\begin{tcolorbox}[colframe=black, colback=white, sharp corners=southwest, boxrule=1pt, width=\textwidth, title = Prompt]

\textbf{Please watch the video clip provided.} After viewing, please estimate the intensity of each listed emotion that people might feel upon viewing the video clip. Rate each emotion on a scale from 0 to 9, where 0 means \textit{`not at all'} and 9 indicates \textit{`very strongly'}.

\bigskip

\textbf{Emotion Categories:} \\
Admiration, Adoration, Aesthetic Appreciation, Amusement, Anger, Anxiety, Awe, Awkwardness, Boredom, Calmness, Confusion, Contempt, Craving, Disappointment, Disgust, Empathic Pain, Entrancement, Envy, Excitement, Fear, Guilt, Horror, Interest, Joy, Nostalgia, Pride, Relief, Romance, Sadness, Satisfaction, Sexual Desire, Surprise, Sympathy, Triumph.

\bigskip

Please rate each emotion individually, following this format:

\texttt{Admiration: [numerical value 0-9] \\
Adoration: [numerical value 0-9] \\
... \\
Triumph: [numerical value 0-9]
}

Respond with numerical values only for each emotion, without additional explanation.

\end{tcolorbox}

\subsection*{Analyzing the commonality of emotion structures}
In this study, Representational Similarity Analysis (RSA) and Gromov–Wasserstein Optimal Transport (GWOT) were both employed to compare emotion structures. RSA is a supervised approach that measures correlations or other metrics between different representational structures based on a predefined label mapping. Its primary advantage lies in the straightforward assessment of the overall similarity in structure. In contrast, GWOT is an unsupervised method that compares structures by automatically searching for optimal correspondences among elements or labels, allowing it to identify flexible relationships without fixing them in advance.

Thus, while RSA excels at quantifying the pure correlation of structural patterns along a fixed mapping, it can overlook relationships that extend beyond the original assumptions. GWOT, on the other hand, derives benefits from adaptively aligning each video's emotional data, making it possible to evaluate precisely which human-reported emotions the model's estimated emotions most closely align with. Specifically, RSA enables evaluation of ``how similar the patterns are'' using measures like correlation coefficients, whereas GWOT measures ``how a model's estimated emotions best match human emotions'' through the optimization of correspondences across the dataset.

Both methods are valuable for comparing emotion structures, yet they differ in their reliance on prior mappings and in their perspectives on similarity. By combining the results from both methods, one can capture different aspects of the emotion structure. In this work, we leveraged these two approaches to offer a multifaceted evaluation of how humans and the model represent emotions when watching videos.

\subsubsection*{Representational Similarity Analysis}
\paragraph{Representation Dissimilarity Matrix (RDM)}
We made a Representation Dissimilarity Matrix (RDM) for the preparation of RSA.
We compiled the emotion-category information for each video into a single ``emotion vector,'' which had 80 dimensions for the Koide-Majima et al. dataset and 34 dimensions for the Cowen \& Keltner dataset. We then computed the cosine similarities between these emotion vectors for different videos and transformed them by \(1 - \text{cosine similarity}\), thereby constructing a Representation Dissimilarity Matrix (RDM) that reflects the differences in emotion ratings across videos.

\paragraph{Correlation between RDMs, RSA}
We evaluated the overall similarity between RDMs by calculating the correlation coefficient between RDMs. This analysis is known as conventional Representational Similarity Analysis (RSA). In this method, we calculated Pearson correlations only from the upper triangular elements of the matrix. It should be noted, however, that this method inherently assumes that emotions induced by the same video are aligned across different similarity structures, which is considered a supervised comparison as opposed to an unsupervised comparison such as Gromov-Wasserstein Optimal Transport, which is discussed below.

\subsubsection*{Gromv-Wasserstein Optimal Transport}

\paragraph{Histogram matching}
Prior to running Gromov-Wasserstein optimal transport (GWOT) we equalized the marginal similarity distributions of the two data sets to remove global scale or bias differences while keeping the procedure fully \emph{unsupervised}.

Let $D\in\mathbb{R}^{n\times n}$ and $D'\in\mathbb{R}^{m\times m}$ be the pairwise‑similarity matrices. Extracting the upper‑triangular elements (excluding the diagonal) gives vectors
\(\mathbf{u}\) and \(\mathbf{v}\) of lengths \(L=n(n-1)/2\) and
\(L'=m(m-1)/2\).  
After sorting both vectors in descending order,
\(\mathbf{u}_{(1)}\ge\ldots\ge\mathbf{u}_{(L)}\) and
\(\mathbf{v}_{(1)}\ge\ldots\ge\mathbf{v}_{(L')}\),
we replace each entry of \(\mathbf{v}\) by the value with the same rank in
\(\mathbf{u}\):
\[
\tilde{\mathbf{v}}_{(r)} \leftarrow \mathbf{u}_{(r)}
\qquad\text{for } r\le\min(L,L').
\]

Because monotone rearrangement minimizes the Wasserstein‑1 distance on
\(\mathbb{R}\), this rank‑wise replacement realizes the optimal
one‑dimensional transport between the two empirical distributions. Only the histogram of similarities is normalized; no pairwise ordering in \(D\) is altered and no cross‑sample correspondences are introduced.  GWOT therefore operates on inputs that share identical marginal statistics, allowing it to focus on higher‑order structural alignment.

\begin{figure}[ht]
    \centering
    \includegraphics[width=\linewidth]{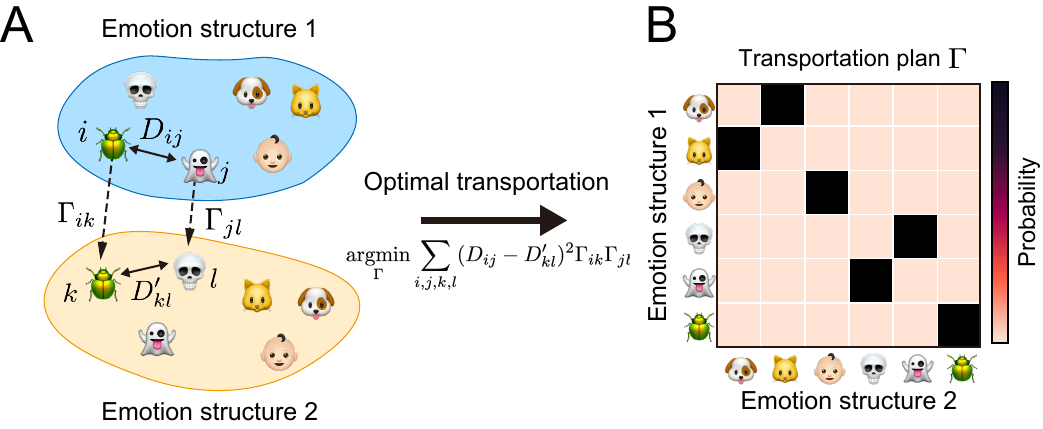}
    \caption{Schematic of the Gromov-Wasserstein optimal transport. A: Each element of $D$ and $D'$ represents the dissimilarity between the emotion ratings of the videos. The optimal transportation plan $\Gamma$ is obtained by minimizing the Gromov-Wasserstein distance (GWD) between the two emotion structures. B: The obtained transportation plan matrix $\Gamma$. Each cell $\Gamma_{ij}$ represents the probability of correspondence between the two videos $i$ and $j$. 
    }
    \label{fig:GWalignment}
\end{figure}

\paragraph{GWOT}
To assess the similarity between emotion structures in an unsupervised method, we applied the Gromov-Wasserstein optimal transport (GWOT) alignment method (Figure \ref{fig:GWalignment}A). GWOT is an unsupervised alignment technique that determine the optimal transportation plan between point sets in two different domains without using information about the correspondence between individual indices. 
In this study, we defined the dissimilarity between video $i$ and $j$ in Emotion structure 1 as $D_{ij}$, and that between video $k$ and $l$ in Emotion structure 2 as $D'_{kl}$, both based on cosine dissimilarity of emotion ratings. To align the structural patterns of these matrices, we obtained the optimal transport plan $\Gamma$ by minimizing the following Gromov-Wasserstein distance (GWD):
\begin{equation}
    \text{GWD} = \min_{\Gamma} \sum_{i,j,k,l} (D_{ij} - D'_{kl})^2 \Gamma_{ik}\Gamma_{jl},
    \label{eq:GWD}
\end{equation}
which quantifies how well the similarity structures of the two domains correspond to each other.

Each element $\Gamma_{ik}$ in optimal transportation plan $\Gamma$ can be interpreted as the probability that the emotional experience elicited by $i$-th video in one domain corresponds to that of $k$-th video in the other domain.
Figure \ref{fig:GWalignment}A illustrates the concept of this optimization. The upper blue region represents emotion structure 1, and the lower orange region represents emotion structure 2. Each point within the structures corresponds to a video, and the arrows indicate the dissimilarity relationships between videos. GWOT identifies the optimal alignment that minimizes the discrepancy in the internal dissimilarity patterns between the two structures.
Figure \ref{fig:GWalignment}B shows the resulting optimal transport plan $\Gamma$, where the rows and columns correspond to videos in emotion structure 1 and 2, respectively. The color represents the transport probability, with darker cells indicating stronger correspondence. For example, the dog video in emotion structure 1 corresponds to the cat video in emotion structure 2, and the baby video corresponds to the baby video across both structures.

Although there is a method to optimize GWD by adding the entropy term, called entropic GWOT, we use GWOT without the entropy term in this paper because we found that GWOT without entropy is faster in terms of computation time and also because the overall performance as evaluated by the matching rate was higher. Note that when we do not add the entropy term, the optimal transportation plan is typically sparse and binary, i.e., each row contains only a single non-zero value in a column. Thus, in this optimization, the optimal transport matrix can be thought of as a permutation matrix, indicating which row (an item in emotion structure 1) corresponds to which column (an item in emotion structure 2). 

To obtain satisfactory local minima, we utilized the GWTune toolbox \cite{Takeda2025-li} to randomly initialize the transport matrix $\Gamma$ and performed multiple optimizations. Since the computational cost of a single optimization increases with the size of the matrix, we adjusted the number of random initializations according to the matrix scale, aiming to balance computational efficiency and alignment accuracy. Specifically, we conducted 10,000 random initializations when the number of videos was approximately 500 or fewer, 1,000 for around 750 videos, and 200 for matrices approaching 2,000 videos. From among the solutions obtained from these trials, we selected the transport plan that minimized the Gromov-Wasserstein distance (GWD).

\subsubsection*{Evaluation of GWOT}
\paragraph{One-to-one matching}
This analysis evaluates (i) how consistently the same video elicits comparable emotion structures across participants and (ii) how closely a model's predicted emotions mirror those structures. For each video, we ask whether the emotion reports from two domains—e.g., Participant group 1 vs.\ Participant group 2, or humans vs.\ the model—refer to the same underlying video. If they do, the match is deemed \emph{correct} and contributes to an overall agreement score.

We compute the correct matching rate based on the optimal transport (OT) plan~$\Gamma$. Let the binary ground truth assignment matrix be $C$, where $C_{ik}=1$ if items~$i$ and~$k$ are a true pair and $0$ otherwise. Using $C$, the OT based matching rate is
\begin{equation}
\text{Matching Rate}
\;=\;
\frac{1}{n}\sum_{i=1}^{n}\sum_{k=1}^{n}
C_{ik}\,
\mathbf{1}\!\bigl(\Gamma_{ik}=\max_{k'}\Gamma_{ik'}\bigr),\label{eq:matching_rate}
\end{equation}
where $\mathbf{1}(\cdot)$ is the indicator function.

Because we use GWOT without the entropy term and the two domains contain the same number of items, $\Gamma$ is a binary permutation matrix. Under this setting, the above expression simplifies to the fraction of non--zero diagonal elements in $\Gamma$; equivalently, it is the percentage of items whose OT mapping lands on the correct counterpart.

\paragraph{Category matching}
To reveal latent common features of emotion structure that cannot be fully captured by strict one-to-one correspondence, we employed a category matching approach. This method relaxes the strict one-to-one evaluation criterion by considering video pairs assigned to the same category as correct matches. Consequently, even if the emotion-report indices do not align perfectly, this approach enables us to assess the similarity among videos that evoke similar emotions.

The category matching rate is computed according to the following procedure. We redefine the ``correct'' assignment matrix \( C \) as follows:
\begin{equation}
C_{ik} =
\begin{cases}
1 & \text{if video } i \text{ and video } k \text{ belong to the same category,} \\
0 & \text{otherwise.}
\label{eq:evaluation_of_category_matching}
\end{cases}
\end{equation}
Subsequently, by performing calculations analogous to those in Equation \eqref{eq:matching_rate}, the category matching rate is obtained.

\paragraph{Determination of video categories}
To evaluate the category matching rate, we derived categories purely from the data because the video stimuli were not annotated with any explicit category labels. A naive approach would be to assign each video to the single emotion with the highest intensity rating; however, we judged that taking the full multivariate pattern of emotion ratings into account would yield a categorization that better reflects the structure of the data. We therefore adopted hierarchical agglomerative clustering — a simple yet standard data‑driven technique — applied to the averaged human emotion‑rating matrix (Ward's linkage, Euclidean distance).

Hierarchical clustering does require pre‑specifying the number of clusters, a choice that is unavoidably somewhat arbitrary. In the present context, however, our aim is merely to construct a category‑level matching metric; any reasonably large number of clusters suffices for this purpose. Pilot analyses across 10–30 clusters produced virtually identical qualitative results, indicating that the precise cut point does not materially affect the downstream matching scores. For the sake of visual clarity and interpretability, we report the 10‑cluster solution in the main text.

Videos grouped into the same cluster are assumed to elicit similar composite emotional responses. These data‑driven clusters serve as the basis for evaluating the commonality of emotion structures both between participants and between participants and models at the categorical level. The category‑level matching rate thus complements stricter one‑to‑one matching measures by capturing broader correspondences in the geometry of emotion space.

\subsection*{Control model of shuffling human's emotion ratings}
In this study, we constructed a shuffled dataset of participants' emotion ratings as a control model to estimate the lower bound of performance. By comparing the estimation accuracy of the actual model to this control model, one can gauge the extent to which the model legitimately captures the video-specific emotion structure. In other words, the shuffled dataset serves as a baseline that reveals how the model would perform if it failed to account for the unique emotional signatures of individual videos, focusing instead on the broad tendencies in the emotion ratings.

Specifically, all reported emotion ratings in the original dataset were randomly permuted, severing the original pairing between each video and its associated ratings. Through this process, the shuffled dataset retains only the marginal distribution of the emotion ratings, thereby approximating a scenario in which the model ``knows the overall distribution of reported emotions, but not their correspondence to specific videos.''

The selection of videos that were ``well estimated'' in each shuffled dataset was performed by ranking the videos according to the correlation between the shuffled human ratings and the original human ratings. Based on these rankings, we extracted various sets of top-performing videos, such as the Top 100, Top 250, and Top 750.
Because the selection is based on correlations with the original human ratings, the top-ranked videos vary across shuffles and may also differ from those selected using the actual model outputs.
Since video selection in all cases is consistently based on the correlation between shuffled and original human ratings, the comparisons remain valid and fair across conditions.

\paragraph{Mean correlation per video and correlation between RDMs}
To estimate the degree of variability introduced by shuffling procedure, we performed 1,000 shuffles of the emotion ratings using different random seeds. 
First, for each video, we calculated the correlation coefficient between the shuffled human ratings and the original human ratings. We then derived the 95\% percentile interval from the distribution of these mean correlation values across videos.
This result is reported in the ``mean of correlation on each video'' row under ``shuffled human ratings'' in Table \ref{tab:nishimoto_Gemini} and Table \ref{tab:CK_table}.
Second, we constructed representational dissimilarity matrices (RDMs) from each of 1,000 shuffled datasets and computed the correlation between each shuffled RDM and the RDM obtained from the original human ratings. The 95\% percentile interval was then derived from the distribution of these correlation values.
This analysis corresponds to the ``correlation of RDMs'' row under ``shuffled human ratings'' in Table \ref{tab:nishimoto_Gemini} and Table \ref{tab:CK_table}.
Through these procedures, we quantitatively assessed the variability of both video-level agreement and structural similarity under random shuffling.

\paragraph{GWOT}
Due to the high computational cost of GWOT analysis, we adopted different procedures depending on the number of videos analyzed. For datasets with approximately 500 videos or fewer, we conducted 10 iterations of shuffling using different random seeds and computed the 95\% percentile interval based on the distribution of the results. In contrast, for datasets with more than 500 videos, we performed only a single shuffle due to the significant time required for computation. This approach allowed us to balance the reliability and feasibility of the GWOT analysis using shuffled data.

\bibliography{paperpile}


\section*{Acknowledgements}
M.O. was supported by JST Moonshot R\&D Grant No. JPMJMS2012 and JSPS KAKENHI, Grant Number 20H05712. M.O. and T.H. were supported by JSPS KAKENHI Grant Number 23H04834. S.N. was supported by JSPS KAKENHI Grant Number JP24H00619.

\section*{Author contributions statement}
H.A., T.H., and M.O. conceptualized the study. H.A., K.N. and M.O. collected data from multimodal LLMs. H.A. performed data analysis. N.K. and S.N. provided experimental data from Koide-Majima et al. (2020) and offered insights regarding data analysis. H.A. and M.O. drafted the initial manuscript. All authors reviewed, edited, and approved the final manuscript.




\end{document}